\documentclass[entropy,article,accept,pdftex,moreauthors]{Definitions/mdpi} 

\firstpage{1} 
\pubvolume{1}
\issuenum{1}
\articlenumber{0}
\pubyear{2023}
\copyrightyear{2023}
\externaleditor{{Academic Editors: Irad E. Ben-Gal and Amichai Painsky}}
\datereceived{1 May 2023} 
\daterevised{24 May 2023} 
\dateaccepted{28 May 2023 } 
\datepublished{ } 
\hreflink{https://doi.org/} 

\usepackage{amsfonts}       
\usepackage{nicefrac}       

\usepackage{pifont}
\newcommand{\cmark}{\ding{51}}%
\newcommand{\xmark}{\ding{55}}%

\usetikzlibrary{bayesnet}
\usetikzlibrary{arrows}

\usepackage{algorithm}

\let\classAND\AND
\let\AND\relax
\usepackage{algorithmic}

\let\AND\classAND
\AtBeginEnvironment{algorithmic}{\let\AND\algoAND}

\usepackage{titletoc}
\usepackage{amsmath}
\usepackage{amssymb}
\usepackage{mathtools}
\usepackage{amsthm}
\usepackage{comment}

\usepackage{cancel} 

\usepackage{amsmath,amsfonts,bm}

\def\eqref#1{(\ref{#1})}

\def\1{\bm{1}}

\def\ry{{\textnormal{y}}}

\def\rvtheta{{\mathbf{\theta}}}

\def\rvx{{\mathbf{x}}}

\def\rvz{{\mathbf{z}}}

\def\rvtheta{{\bm{\theta}}}

\def\vmu{{\bm{\mu}}}
\def\vtheta{{\bm{\theta}}}

\def\vc{{\bm{c}}}

\def\vm{{\bm{m}}}

\def\vw{{\bm{w}}}
\def\vx{{\bm{x}}}
\def\vy{{\bm{y}}}
\def\vz{{\bm{z}}}

\def\mH{{\bm{H}}}

\DeclareMathAlphabet{\mathsfit}{\encodingdefault}{\sfdefault}{m}{sl}
\SetMathAlphabet{\mathsfit}{bold}{\encodingdefault}{\sfdefault}{bx}{n}

\newcommand{\R}{\mathbb{R}}

\Title{On Sequential Bayesian Inference for Continual Learning}

\TitleCitation{On Sequential Bayesian Inference for Continual Learning}

\Author{Samuel Kessler 
 $^{1,}$*, Adam Cobb $^{3}$, Tim G. J. 
Rudner $^{2}$, Stefan Zohren $^{1}$ and Stephen J. Roberts $^{1}$}

\AuthorNames{Samuel Kessler, Adam Cobb, Tim G. J. Rudner, Stefan Zohren and Stephen Roberts}

\AuthorCitation{Kessler, S.; Cobb, A.; Rudner, T.G.J.; Zohren, S.; Roberts, S.J.}

\address{%
$^{1}$ \quad University 
 of Oxford, Department of Engineering Science, Oxford OX2 6ED, UK; \\ 
 skessler@robots.ox.ac.uk (S.K.); zohren@robots.ox.ac.uk (S.Z.); sjrob@robots.ox.ac.uk (S.J.R.)\\
$^{2}$ \quad University of Oxford, Department of Computer Science, Oxford OX1 3QG, UK; tim.rudner@cs.ox.ac.uk (T.G.J.R.)\\
$^{3}$ \quad SRI International, Arlington, VA 22209, US; 
 adam.cobb@sri.com}

\corres{Correspondence:  skessler@robots.ox.ac.uk}

\abstract{
Sequential Bayesian inference can be used for \emph{continual learning} to prevent catastrophic forgetting of past tasks and provide an informative prior when learning new tasks. We revisit sequential Bayesian inference and assess whether using the previous task’s posterior as a prior for a new task can prevent catastrophic forgetting in Bayesian neural networks. Our first contribution is to perform sequential Bayesian inference using Hamiltonian Monte Carlo. We propagate the posterior as a prior for new tasks by approximating the posterior via fitting a density estimator on Hamiltonian Monte Carlo samples. We find that this approach fails to prevent catastrophic forgetting demonstrating the difficulty in performing sequential Bayesian inference in neural networks. Furthermore, we study simple analytical examples of sequential Bayesian inference and CL and highlight the issue of model misspecification which can lead to sub-optimal continual learning performance despite exact inference. Furthermore, we discuss how task data imbalances can cause forgetting. From these limitations, we argue that we need probabilistic models of the continual learning generative process rather than relying on sequential Bayesian inference over Bayesian neural network weights. Our final contribution is to propose a simple baseline called \emph{Prototypical Bayesian Continual Learning}, which is competitive with the best performing Bayesian continual learning methods on class incremental continual learning computer vision benchmarks.
}


\begin{document}


\section{Introduction}
The goal of continual learning (CL) is to find a predictor that learns to solve a sequence of new tasks without losing the ability to solve previously learned tasks. One key challenge of CL with neural networks (NNs) is that model parameters from previously learned tasks are ``overwritten'' during gradient-based learning of new tasks, which leads to \emph{catastrophic forgetting} of previously learned abilities~\citep{mccloskey1989catastrophic, french1999catastrophic}. One approach to CL hinges on using recursive applications of Bayes' Theorem; using the weight posterior in a Bayesian neural network (BNN) as the prior for a new task~\citep{Kirkpatrick}. However, obtaining a full posterior over NN weights is computationally demanding and we often need to resort to approximations, such as the Laplace method~\citep{mackay1992practical} or variational inference~\citep{graves2011practical, blundell2015weight} to obtain a neural network weight posterior.

When performing Bayesian CL, sequential Bayesian inference is performed with an approximate BNN posterior, not the true posterior~\citep{schwarz2018progress, ritter2018online, nguyen2018variational, ebrahimi2019uncertainty, kessler2021hierarchical, loo2020generalized}. If we consider the performance of sequential Bayesian inference with a variational approximation over a BNN weight posterior then we barely observe an improvement over simply learning new tasks with stochastic gradient descent (SGD). We develop this statement further in~\cref{sec:bcl}. So if we had access to the true BNN weight posterior, would this be enough to prevent forgetting by sequential Bayesian inference?

Our contributions in this chapter are to revisit Bayesian CL. 1) Experimentally, we perform sequential Bayesian inference using the true Bayesian NN weight posterior. We do this by using the gold standard of Bayesian inference methods, Hamiltonian Monte Carlo (HMC)~\citep{neal2011mcmc}. We use density estimation over HMC samples and use this approximate posterior density as a prior for the next task within the HMC sampling process. Surprisingly our HMC method for CL yields no noticeable benefits over an approximate inference method (VCL \citep{nguyen2018variational}) despite using samples from the true posterior. 2) As a result we consider a simple analytical example and highlight that exact inference with a misspecified model can still cause forgetting. 3) We show mathematically that under certain assumptions task data imbalances cause forgetting in Bayesian NNs.
4) We propose a new probabilistic model for CL and show that by explicitly modeling the generative process of the data, we can achieve good performance, avoiding the need to rely on recursive Bayesian inference over NN weights to prevent forgetting. Our proposed model, \emph{Prototypical Bayesian Continual Learning} (ProtoCL), is conceptually simple, scalable, and competitive with state-of-the-art Bayesian CL methods in the class-incremental learning setting.

\section{Background}
\unskip
\subsection{The Continual Learning~Problem}
\label{sec:cl}
\emph{Continual learning} (CL) is a learning setting whereby a model must learn to make predictions over a set of tasks sequentially while maintaining performance across all previously learned tasks. In~CL, the~model is sequentially shown $T$ tasks, denoted $\mathcal{T}_{t}$ for $t= 1, \ldots, T$. Each task, $\mathcal{T}_t$, is comprised of a dataset $\mathcal{D}_t = \left \{ (\vx_{i}, y_{i}) \right  \}_{i = 1}^{ N_t} $, which a model needs to learn to make predictions with. More generally, tasks are denoted by distinct tuples comprised of the conditional and marginal data distributions, $\{ p_t(y|\rvx), p_t(\rvx)\}$. After~task $\mathcal{T}_t$, the model will lose access to the training dataset but its performance will be continually evaluated on all tasks $\mathcal{T}_i$ for $i \leq t$. For a thorough overview of different continual learning scenarios, see~\cref{sec:cl_scenarios}.

\subsection{Bayesian Continual~Learning}
\label{sec:bcl}
We consider a setting in which task data arrives sequentially at timesteps, $t = 1, 2, \ldots, T$. At~the first timestep, $t=1$, that is, for task $\mathcal{T}_1$, the model receives the first dataset $\mathcal{D}_1$ and learns the conditional distribution $p(y_i | \vx_i, \rvtheta)$ for all $(\vx_i, y_i) \in \mathcal{D}_1$ ($i$ indexes a datapoint in $\mathcal{D}_1$). We denote the parameters $\rvtheta$ as having a prior distribution $p(\rvtheta)$ for $\mathcal{T}_1$. The~posterior predictive distribution for a test point $\vx_1^* \in \mathcal{D}_1$ is hence:
\begin{align}
\label{eq:pred_dist_t1}
p(y_1^*|\vx_1^*, \mathcal{D}_1) = \int p(y_1^*|\vx_1^*, \rvtheta) p(\rvtheta|\mathcal{D}_1) d \rvtheta.
\end{align}
We note that computing this posterior predictive distribution requires $p(\rvtheta|\mathcal{D}_1)$. For~$t=2$, a~CL model is required to fit $p(y_i | \vx_i, \rvtheta)$ for $(\vx_i, y_i) \in \mathcal{D}_1 \cup \mathcal{D}_2$. The~posterior predictive distribution for a new test point $\vx_2^{*} \in \mathcal{D}_1 \cup \mathcal{D}_2$ point is:
\begin{align}
\label{eq:pred_dist_t2}
p(y_2^*|\vx_2^*, \mathcal{D}_1, \mathcal{D}_2) = \int p(y_2^*|\vx_2^*, \rvtheta) p(\rvtheta|\mathcal{D}_1, \mathcal{D}_2) d \rvtheta.
\end{align}
The posterior must thus be updated to reflect this new conditional distribution. We can use repeated application of Bayes' rule to calculate the posterior distributions $p(\rvtheta|\mathcal{D}_1, \ldots, \mathcal{D}_T)$ as:
\begin{align}
\label{eq:bayes_cl}
    p(\rvtheta|\mathcal{D}_1, \ldots,\mathcal{D}_{T-1}, \mathcal{D}_T) &= \frac{p(\mathcal{D}_T|\rvtheta) p(\rvtheta| \mathcal{D}_1, \ldots,\mathcal{D}_{T-1})}{ p(\mathcal{D}_T | \mathcal{D}_1, \ldots,\mathcal{D}_{T-1})}.
\end{align}

In the CL setting, we lose access to previous training datasets; however, using repeated applications of Bayes' rule~Equation~\eqref{eq:bayes_cl} allows us to sequentially incorporate information from past tasks in the parameters $\rvtheta$. At~$t=1$, we have access to $\mathcal{D}_1$ and the posterior over parameters is:
\begin{align}
    \log p(\rvtheta| \mathcal{D}_1) &= \log p(\mathcal{D}_1|\rvtheta) + \log p(\rvtheta) - \log p(\mathcal{D}_1).
\end{align}
At $t=2$, we require $p(\rvtheta| \mathcal{D}_1, \mathcal{D}_2)$ to calculate the posterior predictive distribution in~Equation~\eqref{eq:pred_dist_t2}. However, we have lost access to $\mathcal{D}_1$. According to Bayes' rule, the~posterior may be written as:
\begin{align}
    \label{eq:bayes_cl_t2}
    \log p(\rvtheta| \mathcal{D}_1, \mathcal{D}_2)
    &=
    \log p(\mathcal{D}_2|\rvtheta) + \log p(\rvtheta|\mathcal{D}_1) - \log p(\mathcal{D}_2|\mathcal{D}_1) ,
\end{align}
where we used the conditional independence of $\mathcal{D}_{2}$ and $\mathcal{D}_{1}$ given $\rvtheta$. We note that the likelihood $p(\mathcal{D}_2 | \rvtheta)$ is only dependent upon the current task dataset, $\mathcal{D}_2$, and~that the prior $p(\rvtheta | \mathcal{D}_1)$ encodes parameter knowledge from the previous task. Hence, we can use the posterior evaluated at $t$ as a prior for learning a new task at $t+1$. From~Equation~\eqref{eq:bayes_cl}, we require that our model with parameters $\rvtheta$ is a sufficient statistic of $\mathcal{D}_1$, i.e.,~$p(\mathcal{D}_2| \rvtheta, \mathcal{D}_1) = p(\mathcal{D}_2 | \rvtheta)$, making the likelihood conditionally independent of $\mathcal{D}_1$ given $\rvtheta$. This observation motivates the use of high-capacity predictors, such as Bayesian neural networks, that are flexible enough to learn from $\mathcal{D}_1$.

\subsubsection*{Continual Learning Example: Split-MNIST}
For the MNIST dataset~\citep{lecun1998gradient} we know that if we were to train a BNN we would achieve good performance by inferring the posterior $p(\rvtheta | \mathcal{D})$~\cref{sec:metrics} and integrating out the posterior to infer the posterior predictive distribution over a test point~\cref{eq:pred_dist_t1}. So if we were to split the dataset MNIST into $5$ two-class classification tasks then we should be able to recursively recover the multi-task posterior $p(\rvtheta | \mathcal{D}) = p(\rvtheta| \mathcal{D}_1 \ldots, \mathcal{D}_5)$ using~\cref{eq:bayes_cl}. This problem is called Split-MNIST \citep{zenke2017continual}, where the first task involves the classification of the digits $\{0, 1\}$ then the second task classification of the digits $\{2, 3\}$ and so on.

We can define three different CL settings~\cite{hsu2018re, van2019three, van2022three}. When we allow the CL agent to make predictions with a task identifier $t$ the scenario is referred to as \emph{task-incremental}. The~identifier $t$ could be used to select different heads~\Cref{sec:cl}, for~instance. This scenario is not compatible with sequential Bayesian inference outlined in~Equation~\eqref{eq:bayes_cl} since no task identifier is required for making predictions. \emph{Domain-incremental} learning is another scenario that does not have access to $t$ during evaluation and requires the CL agent to perform classification to the same output space for each task; for~example, for Split-MNIST the output space is $\{0, 1\}$ for all tasks, so this amounts to classifying between even and odd digits. Domain incremental learning is compatible with sequential Bayesian inference with a Bernoulli likelihood. The~third scenario is \emph{class-incremental} learning which also does not have access to $t$ but the agent needs to classify each example to its corresponding class. For~Split-MNIST, for~example, the~output space is $\{0, \ldots, 9\}$ for each task. Class-incremental learning is compatible with sequential Bayesian inference with a categorical~likelihood.

\subsection{Variational Continual~Learning}
\label{sec:vcl}
Variational CL (VCL;~\citet{nguyen2018variational}) simplifies the Bayesian inference problem in~Equation~\eqref{eq:bayes_cl} into a sequence of approximate Bayesian updates on the distribution over random neural network weights $\bm{\theta}$. To~do so, VCL uses the variational posterior from previous tasks as a prior for new tasks. In~this way, learning to solve the first task entails finding a variational distribution $q_1(\rvtheta|\mathcal{D}_1)$ that maximizes a corresponding variational objective. For~the subsequent task, the~prior is chosen to be $q_1(\rvtheta|\mathcal{D}_1)$, and~the goal becomes to learn a variational distribution $q_2(\rvtheta|\mathcal{D}_2)$ that maximizes a corresponding variational objective under this prior. Denoting the recursive posterior inferred from multiple datasets by $q_t(\rvtheta|\mathcal{D}_{1:t})$, we can express the variational CL objective for the $t$-th task as:
\begin{equation}
\begin{aligned}
    \mathcal{L}(\rvtheta, \mathcal{D}_t) &= \mathbb{D}_{\textrm{KL}} \left[q_t(\rvtheta)||q_{t-1}(\rvtheta|\mathcal{D}_{1:t-1}) \right] - \mathbb{E}_{q_t}[\log p(\mathcal{D}_t | \rvtheta)].
    \label{eq:vcl}
\end{aligned}
\end{equation}

\begin{figure}[H]
\includegraphics[width=0.95\textwidth]{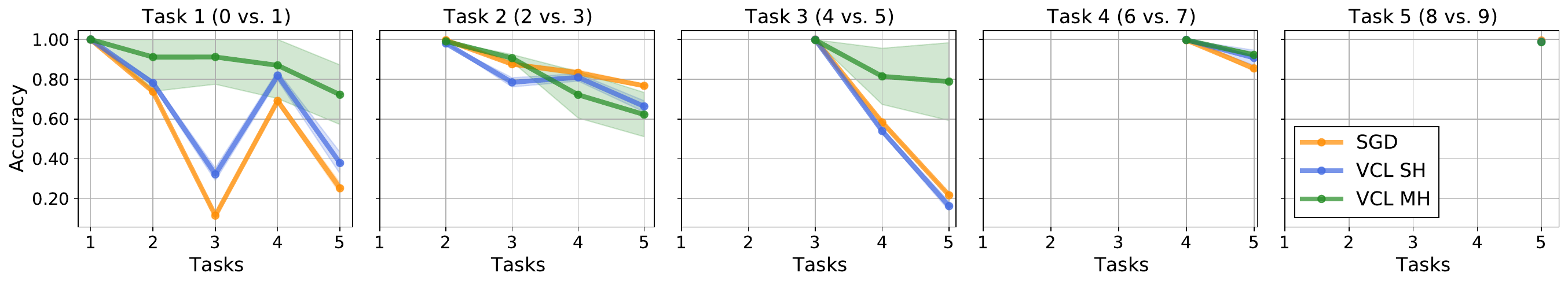}
\caption{Accuracy on Split-MNIST for various CL methods with a two-layer BNN, all accuracies are an average and standard deviation over $10$ runs with different random seeds. We compare an NN trained with SGD (single-headed) with VCL. We consider single-headed (SH) and multi-head (MH) VCL variants, i.e. domain and task incremental learning respectively.}
\label{fig:vcl_accs}
\end{figure}

When applying VCL to the problem of Split-MNIST~Figure~\ref{fig:vcl_accs}, we can see that single-headed VCL barely performs better than SGD when remembering past tasks. Multi-headed VCL performs better, despite not being a requirement from sequential Bayesian inference~Equation~\eqref{eq:bayes_cl}. Therefore, why does single-head VCL not improve over SGD if we can recursively build up an approximate posterior using~Equation~\eqref{eq:bayes_cl}? We hypothesize that it could be due to using a variational approximation of the posterior and so we are not actually strictly performing the Bayesian CL process described in~\Cref{sec:bcl}. We test this hypothesis in the next section by propagating the true BNN posterior to verify whether we can recursively obtain the true multi-task posterior and so improve on single-head VCL and prevent catastrophic~forgetting.

\section{Bayesian Continual Learning with Hamiltonian Monte~Carlo}
\label{sec:hmc_cl}

\begin{figure}[H]
\includegraphics[width=0.6\textwidth]{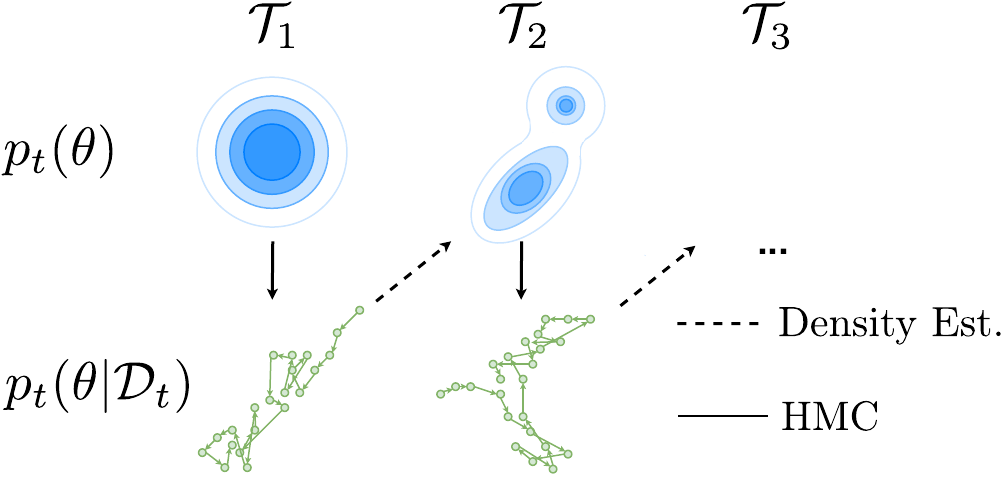}
\caption{Illustration of the posterior propagation process; priors in blue are in the top row and posterior samples on the bottom row. This is a two-step process where we first perform HMC with an isotropic Gaussian prior for $\mathcal{T}_1$ then perform density estimation on the HMC samples from the posterior to obtain $\hat{p}_1(\theta|\mathcal{D}_1)$. This posterior can then be used as a prior for the new task $\mathcal{T}_2$ and so~on.\label{fig:hmc_cl_process}} 
\end{figure}

To perform inference over BNN weights we use the HMC algorithm~\citep{neal2011mcmc}. We then use these samples and learn a density estimator that can be used as a prior for a new task (we considered 
Sequential Monte Carlo, but~it is unable to scale to the dimensions required for the NNs we consider~\citep{chopin2020introduction}. HMC on the other hand has recently been successfully scaled to relatively small BNNs of the size considered in this paper~\citep{cobb2020scaling} and ResNet models but at large computational cost~\citep{izmailov2021bayesian}). HMC is considered the gold standard in approximate inference and is guaranteed to asymptotically produce samples from the true posterior (in the NeurIPS 2021 Bayesian Deep Learning Competition (\url{https://izmailovpavel.github.io/neurips_bdl_competition}), 
the~goal was to find an approximate inference method that is as ``close'' as possible to the posterior samples from HMC). We use posterior samples of $\vtheta$ from HMC and then fit a density estimator over these samples, to~use as a prior for a new task. This allows us to use a multi-modal posterior distribution over $\vtheta$ rather than~a diagonal Gaussian variational posterior such as in VCL. More concretely, to~propagate the posterior $p(\rvtheta | \mathcal{D}_1)$ we use a density estimator, defined $\hat{p}(\rvtheta| \mathcal{D}_{1})$, to~fit a probability density on HMC samples as a posterior. For~the next task $\mathcal{T}_2$ we can use $\hat{p}(\rvtheta| \mathcal{D}_{1})$ as a prior for a new HMC sampling chain and so on (see~Figure~\ref{fig:hmc_cl_process}). The~density estimator priors need to satisfy two key conditions for use within HMC sampling. Firstly, that they are a probability density function. Secondly, that they are differentiable with respect to the input~samples.

\begin{figure}[H]
\begin{adjustwidth}{-\extralength}{0cm}
\includegraphics[width=0.95\linewidth]{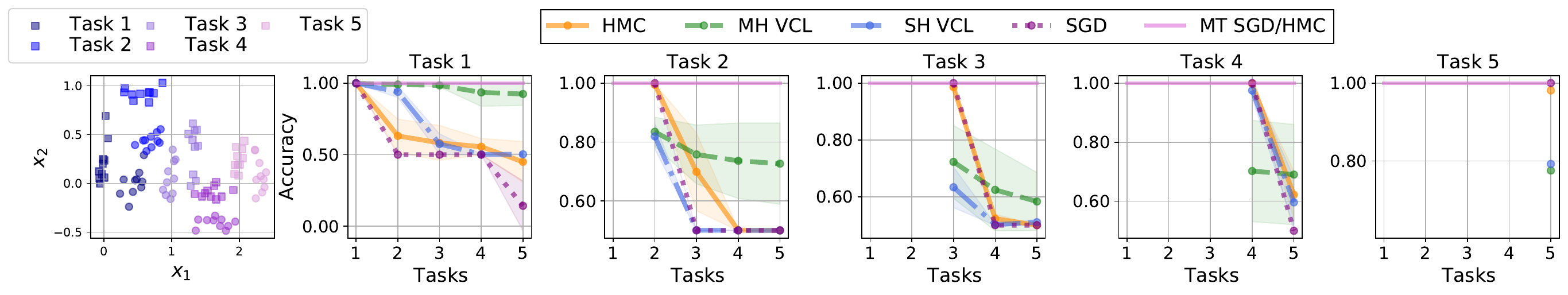}
\end{adjustwidth}
\caption{On the left is the toy dataset of $5$ distinct $2$-way classification tasks that involve classifying circles and squares~\citep{pan2020continual}. 
Moreover, continual learning binary classification test accuracies over $10$ seeds. The~pink solid line is a multi-task (MT) baseline accuracy using SGD/HMC with the same model as for the CL~experiments. This is a domain incremental learning scenario, entirely consistent with sequential Bayesian inference~\cref{sec:metrics}.}
\label{fig:tg_res}
\end{figure}

We use a toy dataset (Figure~\ref{fig:tg_res}) with two classes and inputs $\vx \in \R^2$~\citep{pan2020continual}. Each task is a binary classification problem where the decision boundary extends from left to right for each new task. We train a two-layer BNN, with~a hidden state size of $10$. We use Gaussian Mixture Models (GMM) as a density estimator for approximating the posterior with HMC samples. We also tried Normalizing Flows which should be more flexible~\citep{dinh2016density}; however, these did not work robustly for HMC sampling (RealNVP was very sensitive to the choice of random seed, the~samples from the learned distribution did not give accurate predictions for the current task and led to numerical instabilities when used as a prior within HMC sampling). 
To~the best of our knowledge, we are the first to incorporate flexible priors into the sampling methods such as~HMC.

Training a BNN with HMC on the same multi-task dataset obtains a test accuracy of $1.0$. Thus, the~final posterior is suitable for continual learning under~Equation~\eqref{eq:bayes_cl} and we should be able to recursively arrive at the multi-task posterior with recursive Bayesian inference. This means that if we were to sequentially build up the posterior~\cref{eq:bayes_cl} then we should expect an accuracy of $1.0$ as well since the multi-task posterior predictive performance is an upper bound to the performance of the sequential Bayesian inference for continual learning~\cref{sec:metrics}.

The~results from Figure~\ref{fig:tg_res} demonstrate that using HMC with an approximate multi-modal posterior fails to prevent forgetting and is less effective than using multi-head VCL. In~fact, multi-head VCL clearly outperforms HMC, indicating that the source of the knowledge retention is not through the propagation of the posterior but through the task-specific heads. For~$\mathcal{T}_{2}$, we use $\hat{p}(\rvtheta | \mathcal{D}_1)$ instead of $p(\rvtheta | \mathcal{D}_1)$ as a prior and this will bias the HMC sampling for all subsequent tasks. In~the next paragraph, we detail the measures taken to ensure that our HMC chains have converged so we can assume that we are sampling from the true posterior. Moreover, we assess the fidelity of the GMM density estimator with respect to the HMC samples. We also repeated these experiments with another toy dataset of five binary classification tasks where we observed similar results~\Cref{sec:toy_gaussians}.

For HMC, we ensure that we are sampling from the posterior by assessing chain convergence and effective sample sizes~(Figure~\ref{fig:td_convergence_diagnostics}). The~effective sample size measures the autocorrelation in the chain. The~effective sample sizes for the HMC chains for our BNNs are similar to the literature~\citep{cobb2020scaling}. Moreover, we ensure that the GMM approximate posterior is multi-modal and has a more complex posterior in comparison to VCL, and~that the GMM samples produce equivalent results to HMC samples for the current task~(Figure~\ref{fig:td_gmm_diagnostics}). See~Appendix~\ref{sec:hmc_impl_details} 
 for~details.

The $2$-d benchmarks we consider in this section are from previous works and are domain-incremental continual learning problems. The~domain incremental setting is also simpler~\citep{van2022three} than the class-incremental setting and thus a good starting point when attempting to perform exact sequential Bayesian inference. Despite this, we are not able to perform sequential Bayesian inference in BNNs despite using HMC, which is considered the gold standard of Bayesian deep learning. HMC and density estimation with a GMM produces richer, more accurate, and~multi-modal posteriors. Despite this, we are still not able to sequentially build up the multi-task posterior or obtain much better results than an isotropic Gaussian posterior such as single-head VCL. The~weak point of this method is the density estimation, the~GMM removes probability mass over areas of the BNN weight space posterior, which is important for the new task. This demonstrates just how difficult a task it is to model BNN weight posteriors. In~the next section, we study a different analytical example of sequential Bayesian inference and look at how model misspecification and task data imbalances can cause forgetting in Bayesian~CL.

\section{Bayesian Continual Learning and Model Misspecification}
\label{sec:misspecification}

\begin{figure}
    \includegraphics[width=0.7\textwidth]{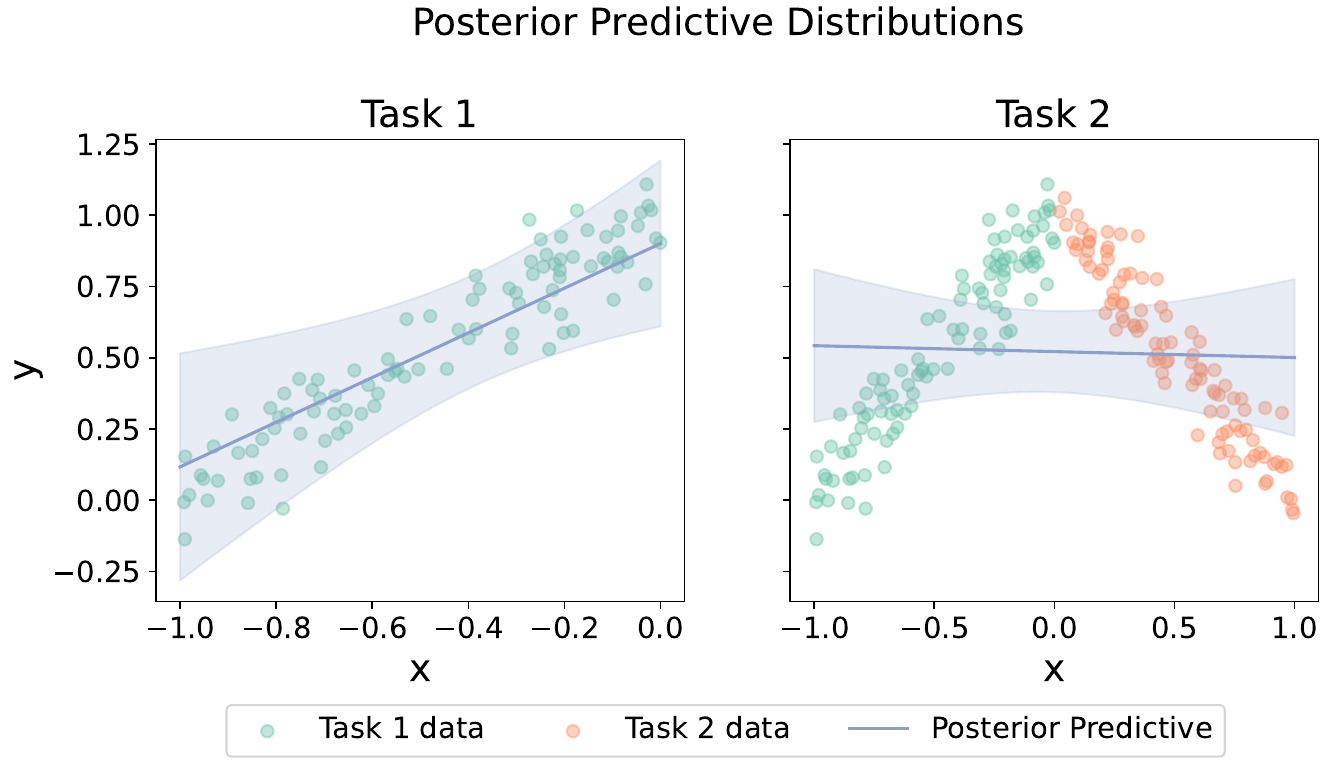}
    \caption{Posterior predictive distributions for a Bayesian linear regression model. Left, data comes from a single task that the Bayesian linear model can fit well. Right, a new dataset is obtained from a different part of the domain, and our sequentially updated Bayesian linear regression models (correctly under Bayesian inference) a global solution to both of these datasets which is sub-optimal for both.} 
\label{fig:bayes4cl:changepoint}
\end{figure}

We now consider a simple analytical example where we can perform the sequential Bayesian inference~\cref{eq:bayes_cl} in closed form using conjugacy. We consider a Bayesian linear regression model with Gaussian likelihoods for $2$ continual learning tasks. This simple example highlights that forgetting~\Cref{eq:metrics:forgetting} may occur under certain conditions despite correct inference.

We use a Gaussian likelihood of the form $p(\mathcal{D}|\vtheta) = \mathcal{N}(y; f(X;\vtheta), \beta^{-1})$ such that $y = f(X ; \vtheta) + \epsilon$ where $\epsilon \sim \mathcal{N}(0, \beta^{-1})$ and $f(X; \vtheta) = \vtheta X^{\top}$. We put a Gaussian prior over the parameters $\vtheta$ such that $p(\vtheta) = \mathcal{N}(\vtheta; \vm_0, \Sigma_0)$ for our first task. Via conjugacy of the Gaussian prior and likelihood, the posterior is also Gaussian, $p(\vtheta | \mathcal{D}) = \mathcal{N}(\vtheta; \vm, \Sigma)$ where $\vm = \Sigma(\Sigma_0^{-1}\vm_0 + \beta X^{\top} \vy)$ and $\Sigma^{-1} = \Sigma_0^{-1} + \beta X^{\top}X$ for task $2$ and onwards. By using sequential Bayesian inference we can have closed-form update equations for our parameters.

From the form of the linear regression posterior, our model can only model linear data. So, if we have data that is linear and drawn from a second task, from a distinct part of the domain, then the model correctly models a linear model, which is the Bayes solution of the multi-task problem~\cref{sec:metrics}. For example, the task $1$ dataset is generated according to $y = x + 1 + \epsilon$, where $\epsilon = \mathcal{N}(0, \beta^{-1})$ for $x \in [-1, 0)$. From~\cref{fig:bayes4cl:changepoint} we can see that our Bayesian linear regression accurately models this first dataset. Now, if we sequentially model a second dataset, with data drawn from $y = -x + 1 + \epsilon$, where $\epsilon = \mathcal{N}(0, \beta^{-1})$ for $x \in [0, 1]$. The model regresses to both of these datasets: the continual learning regression from using the posterior from task $1$ as a prior for task $2$ is the same as if we were to regress to the multi-task dataset of both tasks $1$ and $2$ (see~\cref{fig:bayes4cl:changepoint} on the right)\footnote{A previous version of this section had a similar example where data for both tasks was generated from the same domains $p_1(\vx) = p_2(\vx)$, but the conditional distributions mapped to different outputs $p_1(y|\vx) \neq p_{2}(y|\vx)$. In this previous example, the tasks were conflicting and so the Bayesian multitask posterior was also sub-optimal for continual learning. Now, in this $2$ task linear regression setup, the domains for both tasks are distinct and so the tasks are not conflicting.}.

However, as we can see from~\cref{fig:bayes4cl:changepoint} (right), this is a suboptimal continual learning solution, since we want the average performance~\cref{eq:metrics:av_performance} to be high for all tasks. More specifically the performance after learning task $2$ is $P_2 = p_{2, 2} + p_{2, 1}$ where $p_{i, j}$ is the performance for task $j$ after learning task $i$ and $j \leq i$. Higher is better for the performance measure $p_{i, j}$, for regression $p_{i, j}$ could be the log-likelihood. We see however, that performance is low for the exact sequential Bayesian linear regression continual learning model~\cref{fig:bayes4cl:changepoint}. As with all continual learning benchmarks, we require our model to perform equally well on both tasks. In this case, we can specify a better model which is not just a Bayesian linear regression model, but a mixture of linear regressors~\cite{bishop2006pattern}. \emph{Despite performing exact inference a misspecified model can forget}. We get forgetting of task $1$ after learning task $2$, since the forgetting metric~\cref{eq:metrics:forgetting}, is $f_{1}^2 = p_{1, 1} - p_{2, 1} > 0$ since $p_{1, 1} > p_{2,1}$ which indicates forgetting, as can be seen from~\cref{fig:bayes4cl:changepoint}.

In the case of HMC, we verified that our Bayesian neural network had perfect performance on all tasks \emph{beforehand}. In~\cref{sec:hmc_cl} we had a well-specified model but struggled with exact sequential Bayesian inference~\cref{eq:bayes_cl}. With this Bayesian linear regressions scenario we are performing exact inference, however, we have a misspecified model and so unable to obtain good performance on both tasks~\cref{sec:metrics}. It is important to disentangle model misspecification and exact inference and highlight that model misspecification is a caveat that has not been highlighted in the CL literature as far as we are aware. Furthermore, we can only ensure that our models are well specified if we have access to data from all tasks a priori. So in the scenario of \emph{online continual learning}~\citep{aljundi2019gradient, aljundi2019task, de2019continualarxiv} we cannot know if our model will perform well on all past and future tasks without making assumptions on the task distributions.

\section{Sequential Bayesian Inference and Imbalanced Task~Data}
\label{sec:data_imbalance}

Neural Networks are complex models with a broad hypothesis space and hence are a suitably well-specified model when tackling continual learning problems~\citep{wilson2020bayesian}. However, we struggle to fit the posterior samples from HMC to perform sequential Bayesian inference in~\Cref{sec:hmc_cl}. 

\begin{figure}[H]
     \resizebox{0.75\linewidth}{!}{
     \begin{tikzpicture}
      \node[latent]                            (wtm1) {$\theta^*_{t-1}$};
      \node[latent, left=of wtm1]              (wtm2) {$\theta^*_{t-2}$};
      \node[latent, right=of wtm1]              (wt) {$\theta^*_{t}$};
      \node[latent, right=of wt]                (wtp1) {$\theta^*_{t+1}$};
      \node[obs, above=of wtm1]                 (ytm1) {$y_{t-1}$};
      \node[obs, above=of wtm2]                 (ytm2) {$y_{t-2}$};
      \node[obs, above=of wt, yshift=0.15cm]                 (yt) {$y_{t}$};
      \node[obs, above=of wtp1]               (ytp1) {$y_{t+1}$};
      \node[obs, below=of wtm1]  (xtm1) {$x_{t-1}$};
      \node[obs, below=of wtm2]  (xtm2) {$x_{t-2}$};
      \node[obs, below=of wt, yshift=-0.15cm]    (xt) {$x_{t}$};
      \node[obs, below=of wtp1]  (xtp1) {$x_{t+1}$};
      
      \node[const, left=of wtm2]  (wtm3)  {$\, \ldots \,$}; %
      \node[const, right=of wtp1]  (wtp2)  {$\, \ldots \,$}; %
      
      \edge {wtm3} {wtm2} ; %
      \edge {wtm2} {wtm1} ; %
      \edge {wtm1} {wt} ; %
      \edge {wt} {wtp1} ; %
      \edge {wtp1} {wtp2} ; %
      
      \edge {wtm2} {ytm2} ; %
      \edge {wtm1} {ytm1} ; %
      \edge {wt} {yt} ; %
      \edge {wtp1} {ytp1} ; %
      
      \path (xtm2) edge [bend right, ->]  (ytm2) ;
      \path (xtm1) edge [bend right, ->]  (ytm1) ;
      \path (xt) edge [bend right, ->]  (yt) ;
      \path (xtp1) edge [bend right, ->]  (ytp1) ;

      
    \end{tikzpicture}
    }
     \caption{Graphical model for filtering. Grey and white nodes and latent variables are observed,~respectively.} 
\label{fig:bnn_kf_gm_main}
\end{figure}
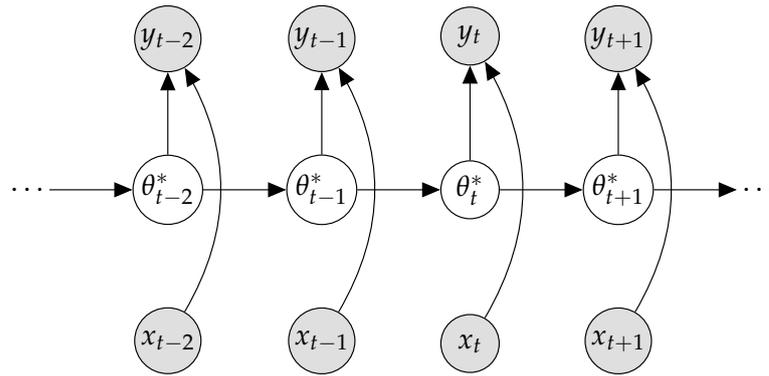

We continue to use Bayesian filtering and assume a Bayesian NN where the posterior is Gaussian with a full covariance. By~modeling the entire covariance, we enable modeling of how each individual weight varies with respect to all others. We do this by interpreting online learning in Bayesian NNs as filtering~\citep{ciftcioglu1995adaptive}. Our treatment is similar to~\citet{aitchison2020bayesian}, who derives an optimizer by leveraging Bayesian filtering. We consider inference in the graphical model depicted in~Figure~\ref{fig:bnn_kf_gm_main}. The~aim is to infer the optimal BNN weights, $\rvtheta^*_t$ at $t$ given a single observation and the BNN weight prior. The~previous BNN weights are used as a prior for inferring the posterior BNN parameters. We consider the online setting, where a single data point $(\vx_t, y_t)$ is observed at a~time.

Instead of modeling the full covariance, we instead consider each parameter $\theta_i$ as a function of all the other parameters $\rvtheta_{-it}$. 
We also assume that the values of the weights are close to those of the previous timestep~\citep{jacot2018neural}. To~obtain the updated equations for BNN parameters given a new observation and prior, we make two simplifying assumptions as~follows.

\begin{Assumption}
\label{assumption:1}
For a Bayesian neural network with output $f(\vx_t; \rvtheta)$ and likelihood $\mathcal{L}(\vx_t, y_t; \rvtheta)$, the~derivative evaluated at $\rvtheta_t$ is $\rvz_t = \partial \mathcal{L}(\vx_t, y_t;\rvtheta) / \partial \rvtheta |_{\rvtheta = \rvtheta_t}$ and the Hessian is $\mH$. We assume a quadratic loss for a data point $(\vx_t, y_t)$ of the form:
\begin{align}
\SwapAboveDisplaySkip
    \mathcal{L}(\vx_t, y_t; \rvtheta) = \mathcal{L}_t(\rvtheta) = -\frac{1}{2}\rvtheta^{\top} \mH \rvtheta + \rvz^{\top}_{t} \rvtheta,
\end{align} 
the result of a second-order Taylor expansion. The~Hessian is assumed to be constant with respect to $(\vx_t, y_t)$ (but not with respect to $\vtheta$).
\end{Assumption}
To construct the dynamical equation for $\rvtheta$, consider the gradient for the $i$-th weight while all other parameters are set to their current estimate at the optimal value for the $\theta_{it}^{*}$:
\begin{align}
    \label{eq:bnn_kf_dynamics}
    \theta_{it}^{*} = -\frac{1}{H_{ii}} \mH_{-ii}^{\top} \rvtheta_{-it},
\end{align}
since $z_{it} = 0$ at a mode. The~equation above shows us that the dynamics of the optimal weight $\theta^{*}_{it}$ is dependent on all the other current values of the parameters $\rvtheta_{-it}$. The~dynamics of $\rvtheta_{-it}$ are a complex stochastic process dependent on many different variables such as the dataset, model architecture, learning rate schedule, etc.
\begin{Assumption}
\label{assumption:2}
Since reasoning about the dynamics of $\rvtheta_{-it}$ is intractable, we assume that at the next timestep, the optimal weights are close to the previous timesteps with a discretized Ornstein--Uhlenbeck process for the weights $\rvtheta_{-it}$ with reversion speed $\vartheta \in \R_{+}$ and noise variance $\eta^2_{-i}$:
\begin{align}
    p(\rvtheta_{-i,t+1} | \rvtheta_{-i ,t}) = \mathcal{N}((1-\vartheta)\rvtheta_{-it}, \eta_{-i}^2),
\end{align}
this implies that the dynamics for the optimal weight are defined by
\begin{align}
\label{eq:transition_dynamics}
    p(\theta^*_{i,t+1} | \theta^*_{i,t}) = \mathcal{N}((1-\vartheta) \theta^*_{it}, \eta^2),
\end{align}
where $\eta^2 = \eta^2_{-i} \mH^{\top}_{-ii} \mH_{-ii}$.
\end{Assumption}
In simple terms, in~Assumption~\ref{assumption:2}, we assume a parsimonious model of the dynamics, and that the next value of $\vtheta_{-i, t}$ is close to their previous value according to a Gaussian, similarly to~\citet{aitchison2020bayesian}.
\begin{Lemma} 
\label{lemma:1}
\textit{Under Assumptions~\ref{assumption:1} and \ref{assumption:2} the dynamics and likelihood are Gaussian. Thus, we are able to infer the posterior distribution over the optimal weights using Bayesian updates and by linearizing the BNN the update equations for the posterior of the mean and variance of the BNN for a new data point are:}
\begin{align}
    \mu_{t, \text{post}}
    &=
    \sigma^2_{t, \text{post}} \left(\frac{\mu_{t, \text{prior}}}{\sigma^2_{t, \text{prior}}(\eta^2)} + \frac{y_t}{\sigma^2}g(\vx_t)\right) \qquad \text{and} \qquad
    \frac{1}{\sigma^2_{t, \textrm{post}}} =
    \frac{g(\vx_t)^2}{\sigma^2} + \frac{1}{\sigma^2_{t, \textrm{prior}}(\eta^2)}
    \label{eq:bnn_kf} ,
\end{align}
\textit{where we drop the notation for the $i$-th parameter, the~posterior is $\mathcal{N}(\theta^*_{t}; \mu_{t, \text{post}}, \sigma^2_{t, \text{post}})$ and $g(\vx_t) = \frac{\partial f(\vx_t; \theta^{*}_{it})}{\partial \theta^*_{it}}$ and $\sigma^2_{t, \textrm{prior}}$ is a function of $\eta^2$.}
\end{Lemma}

See~Appendix \ref{sec:bayes_as_kf_appendix} for the derivation of Lemma~\ref{lemma:1}. From~Equation~\eqref{eq:bnn_kf}, we can notice that the posterior mean depends linearly on the prior and a data-dependent term and so will behave similarly to our previous example in~\Cref{sec:misspecification}. Under~Assumption~\ref{assumption:1} and Assumption~\ref{assumption:2}, if there is a data imbalance between tasks in~Equation~\eqref{eq:bnn_kf}, then the data-dependent term will dominate the prior term if there is more data for the current~task.

In~\Cref{sec:hmc_cl}, we showed that it is very difficult with current machine learning tools to perform sequential Bayesian inference for simple CL problems with small Bayesian NNs. When we disentangle Bayesian inference and model misspecification, we show showed that misspecified models can forget despite exact Bayesian inference. The~only way to ensure that our model is well specified is to show that the multi-task posterior produces reasonable posterior predictive distributions $p(y|\vx, \mathcal{D}) = \int p(y | \vx, \mathcal{D}, \vtheta) p(\vtheta | \mathcal{D}) d\vtheta$ for one's application. Additionally, in~this section, we have shown that if there is a task dataset size imbalance, then we can obtain forgetting under certain~assumptions.

\section{Related~Work}

There has been a recent resurgence in the field of CL~\citep{thrun1995lifelong} given the advent of deep learning. Methods that approximate sequential Bayesian inference~Equation~\eqref{eq:bayes_cl} have been seminal in CL's revival and have used a diagonal Laplace approximation~\citep{Kirkpatrick, schwarz2018progress}. The~diagonal Laplace approximation has been enhanced by modeling covariances between neural network weights in the same layer~\citep{ritter2018online}. Instead of the Laplace approximation, we can use a variational approximation for sequential Bayesian inference, named VCL~\citep{nguyen2018variational, zeno2018task}. The variational Gaussian variance of each Bayesian NN parameter can be used to pre-condition the learning rates of each weight and create a mask per task by using pruning~\cite{ebrahimi2019uncertainty}. Using richer priors has also been explored~\citep{ahn2019uncertainty, farquhar2020radial, kessler2021hierarchical, mehta2021continual, kumar2021bayesian}. For example, one can learn a scaling of the Gaussian NN weight parameters for each task by learning a new variational adaptation parameter which can strengthen the contribution of a specific neuron~\cite{adel2019continual}. The~online Laplace approximation can be seen as a special case of VCL where the KL-divergence term~Equation~\eqref{eq:vcl} is tempered and the temperature tends to $0$~\cite{loo2020generalized}. Gaussian processes have also been applied to CL problems leveraging inducing points to retain previous task functions~\citep{titsias2020functional, kapoor2021variational}.

Bayesian methods that regularize weights have not matched up to the performance of experience replay-based CL methods~\citep{buzzega2020dark} in terms of accuracy on CL image classification benchmarks. Instead of regularizing high-dimensional weight spaces, regularizing task functions is a more direct approach to combat forgetting~\citep{benjamin2018measuring}. Bayesian NN weights can also be generated by a hypernetwork, where the hypernetwork needs only simple CL techniques to prevent forgetting~\citep{henning2021posterior}. In~particular, one can leverage the duality between the Laplace approximation and Gaussian processes to develop a functional regularization approach to Bayesian CL~\citep{swaroop2019improving} or using function-space variational inference~\citep{Rudner2022fsvi, Rudner2021cfsvi}.

In the next section, we propose a simple Bayesian continual learning baseline that models the data-generating continual learning process and performs exact sequential Bayesian inference in a low-dimensional embedding space. Previous work has explored modeling the data-generating process by inferring the joint distribution of inputs and targets $p(\rvx, \ry)$ and learning a generative model to replay data to prevent forgetting~\citep{lavda2018continual}, and~by learning a generative model per class and evaluating the likelihood of the inputs given each class $p(\rvx | \ry)$~\citep{van2021class}.


\section{Prototypical Bayesian Continual~Learning}
\label{sec:proto_cl}

We have shown that sequential Bayes over NN parameters is very difficult~(\cref{sec:hmc_cl}), and is only suitable for situations where the multi-task posterior is suitable for all tasks. We now show that a more fruitful approach is to model the generative CL problem with a generative classifier where each class is represented by a prototype and classification is distance based to the prototype. This is simple and scalable. In particular, we represent classes by prototypes~\citep{snell2017prototypical, rebuffi2017icarl} and maintain prototypes with a replay buffer to prevent catastrophic forgetting.
We refer to this framework as Prototypical Bayesian Continual Learning, or ProtoCL for short. This approach can be viewed as a probabilistic variant of iCarl~\citep{rebuffi2017icarl}, which creates embedding functions for different classes which are simply class means and predictions are made by nearest neighbors. ProtoCL also bears similarities to the few-shot learning model Probabilistic Clustering for Online Classification~\citep{harrison2020continuous} and MetaQDA~\cite{zhang2021shallow}, developed for few-shot image classification. MetaQDA uses Normal-inverse Wishart conjugate priors for Gaussian quadratic discriminant analysis (QDA), ProtoCL has different design choices as follows.

\begin{figure}[H]
    \includegraphics[width=0.50\textwidth, keepaspectratio,trim={80pt 0 40pt 0},clip]{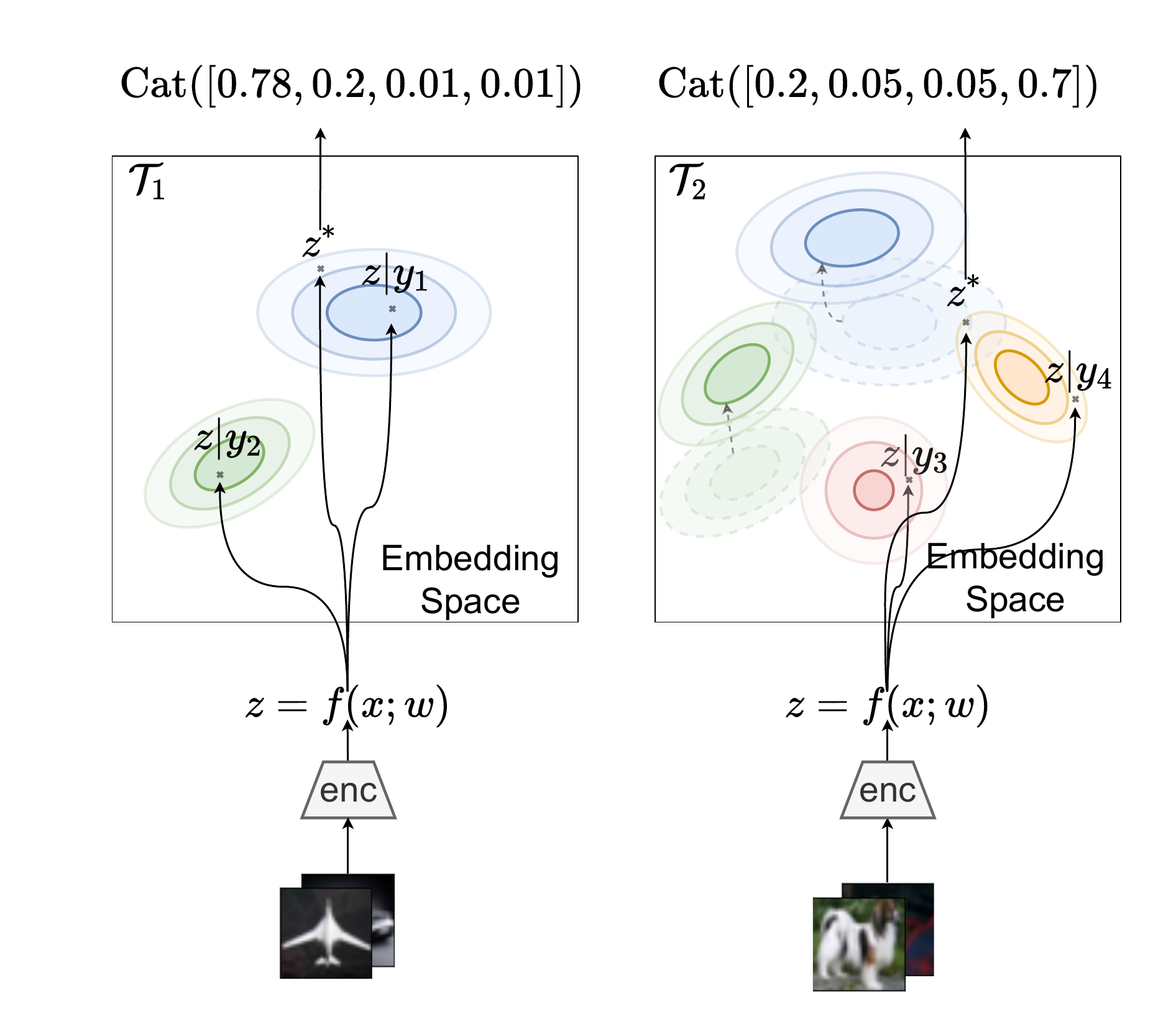}
    \caption{Overview of~ProtoCL.}
    \label{fig:protocl_overview}
\end{figure}

\textbf{Model.} ProtoCL models the generative CL process. We consider classes $j \in \{1, \ldots, J\}$, generated from a categorical distribution with a Dirichlet prior:
\begin{align}
    y_{i, t} \sim \text{Cat}(p_{1:J}), \quad p_{1:J}\sim \textrm{Dir}(\alpha_t).
\end{align} 
Images are embedded into an embedding space by an encoder, $\vz = f(\vx; \vw)$ with parameters $\vw$. The per-class embeddings are Gaussian whose with isotropic variance. The prototype mean has a prior which is also Gaussian with and diagonal covariance:
\begin{align}
    \vz_{it}|y_{it} \sim \mathcal{N}(\bar{\vz}_{yt}, \Sigma_{\epsilon}), \quad \bar{\vz}_{yt} \sim \mathcal{N}(\vmu_{yt}, \Lambda^{-1}_{yt}).
\end{align}

See~\cref{fig:protocl_overview} for an overview of the model. To alleviate forgetting in CL, ProtoCL uses a coreset of past task data to continue to embed past classes distinctly as prototypes. The posterior distribution over class probabilities $\{p_j\}_{j=1}^J$ and class embeddings $\{\bar{z}_{y_j}\}_{j=1}^J$ is denoted in short hand as $p(\vtheta)$ with parameters $\eta_{t} = \{\alpha_{t}, \vmu_{1:J, t}, \Lambda^{-1}_{1:J, t}\}$. We model the Gaussians with a diagonal covariance. ProtoCL models each class prototype but does not use task-specific NN parameters or modules like multi-head VCL. ProtoCL uses a probabilistic model over an embedding space which allows it to use powerful embedding functions $f(\, \cdot \,; \vw)$ without having to parameterize them probabilistically and so this approach will be more scalable than VCL, for instance.

\noindent \textbf{Inference.} As the Dirichlet prior is conjugate with the Categorical distribution and likewise the Gaussian over prototypes with a Gaussian prior over the prototype mean, we can calculate posteriors in closed form and update the parameters $\eta_t$ as new data is observed without using gradient-based updates. We optimize the model by maximizing the posterior predictive distribution and use a softmax over class probabilities to perform predictions. We perform gradient-based learning of the NN embedding function $f(\,\cdot\,; \vw)$ and update the parameters, $\eta_t$ at each iteration of gradient descent as well, see~\cref{alg:protocl}. 

\noindent \textbf{Sequential updates.} We can obtain our parameter updates for the Dirichlet posterior by Categorical-Dirichlet conjugacy:
\begin{align}
\SwapAboveDisplaySkip  
    \alpha_{t+1, j} = \alpha_{t, j} + \sum^{N_t}_{i=1}\mathbb{I}(y^i_{t} = j), \label{eq:proto_cl_dir_updates}
\end{align}
where $N_t$ are the number of points seen during the update at time step $t$. Also, due to Gaussian-Gaussian conjugacy, the posterior for the Gaussian prototypes is governed by:
\begin{align}
    \Lambda_{y_{t+1}} &= \Lambda_{y_t} + N_{y} \Sigma^{-1}_{\epsilon} \label{eq:proto_cl_norm_cov_update}\\
    \Lambda_{y_{t+1}}\vmu_{y_{t+1}} &= N_{y} \Sigma^{-1}_{\epsilon} \bar{\vz}_{y_t} + \Lambda_{y_t} \vmu_{y_t}, \, \forall y_t \in C_t, \label{eq:proto_cl_norm_mean_update}
\end{align}
where $N_{y}$ are the number of samples of class $y$ and $\bar{\vz}_{y_t} = (1 / N_{y}) \sum^{N_y}_{i=1} z_{yi}$, see~\cref{sec:protocl_derivation} for the detailed derivation.

\noindent \textbf{Objective.} We optimize the posterior predictive distribution of the prototypes and classes:
\begin{align}
    p(\rvz, y) &= \int p(\rvz, y| \vtheta_t ; \eta_t)p(\vtheta_t ; \eta_t)d\vtheta_t
    = p(y) \prod^{N_t}_{i=1} \mathcal{N}(\rvz_{it} | y_{it} ; \vmu_{y_t, t}, \Sigma_{\epsilon} + \Lambda^{-1}_{y_t, t}) \label{eq:proto_cl_log_post}.
\end{align}

Where the $p(y) = \alpha_y / \sum^{J}_{j=1} \alpha_j$, see~\cref{sec:protocl_obj} for the detailed derivation. This objective can then be optimized using gradient-based optimization for learning the prototype embedding function $\vz=f(\vx;\vw)$.

\noindent\textbf{Predictions.} To make a prediction for a test point $\vx^*$ the class with the maximum (log)-posterior predictive is chosen, where the posterior predictive is:
\begin{align}
    p(y^*=j | \vx^*, \vx_{1:t}, y_{1:t}) &= p(y^*=j | \vz^*, \vtheta_t)
    =
    \frac{p(y^* =j , \vz^* | \vtheta_t)}{\sum_{i}p(y=i, \vz^* | \vtheta_t)},
\end{align}
see~\cref{sec:proto_cl_predictions} for further details.

\begin{algorithm}[t!]
   \caption{ProtoCL continual learning}
   \label{alg:protocl}
    \begin{algorithmic}[1]
   \STATE {\bfseries Input:} task datasets $\mathcal{T}_{1:T}$
, initialize embedding function: $f(\, \cdot \,; \vw)$, coreset: $\mathcal{M} = \emptyset$.   
   \FOR{$\mathcal{T}_1$ {\bfseries to} $\mathcal{T}_T$}
   \FOR{each batch in $\mathcal{T}_i \cup \mathcal{M}$}
   \STATE Optimize $f(\cdot; \vw)$ by maximizing the posterior predictive $p(\rvz, y)$~\cref{eq:proto_cl_log_post}
   \STATE Obtain posterior over $\rvtheta$ by updating $\eta$,~\cref{eq:proto_cl_dir_updates,eq:proto_cl_norm_cov_update,eq:proto_cl_norm_mean_update}.
   \ENDFOR
   \STATE Add random subset from $\mathcal{T}_i$ to $\mathcal{M}$.
   \ENDFOR
\end{algorithmic}
\end{algorithm}

\setlength{\tabcolsep}{20.5pt}
\begin{table*}[b!]
  \centering
  \caption{Mean accuracies across all tasks over CL vision benchmarks for \emph{class incremental learning}~\citep{van2019three}. All results are averages and standard errors over $10$ seeds. $^{*}$Uses the predictive entropy to make a decision about which head for class incremental learning.}
  \label{tab:bayes4cl:proto_cl_res}
   \resizebox{1.0\linewidth}{!}{
  \begin{tabular}{lccc}
  \toprule
  \bfseries Method & \bfseries Coreset & \bfseries Split-MNIST & \bfseries Split-FMNIST  \\
  \midrule
  VCL~\citep{nguyen2018variational} & \xmark & $33.01 \pm 0.08$ & $32.77 \pm 1.25$ \\
   $\quad \quad +$ coreset & \cmark & \hspace*{5pt}$52.98 \pm 18.56$ & \hspace*{5pt}$61.12 \pm 16.96$ \\
  HIBNN$^{*}$~\citep{kessler2021hierarchical} & \xmark & {$85.50 \pm 3.20$} & \hspace*{5pt}$43.70 \pm 20.21$  \\
  FROMP~\citep{pan2020continual} & \cmark & {$84.40 \pm 0.00$} & $68.54 \pm 0.00$  \\
  S-FSVI~\citep{Rudner2021cfsvi} & \cmark & $\bm{92.94 \pm 0.17}$ & $80.55 \pm 0.41$ \\
  ProtoCL (\textbf{ours}) & \cmark & $\bm{93.73 \pm 1.05}$ & $\bm{82.73 \pm 1.70}$ \\
  \bottomrule
  \end{tabular}
  }
\end{table*}

\begin{figure}
    \includegraphics[width=0.5\textwidth]{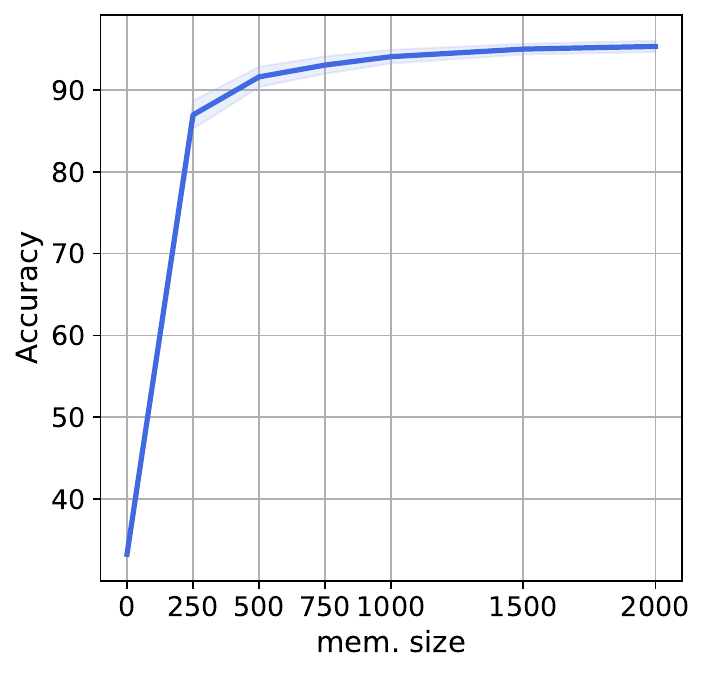}
    \caption{Split-MNIST average test accuracy over $5$ tasks for different memory sizes. On the x-axis we show the size of the entire memory buffer shared by all $5$ tasks. Accuracies are over a mean and standard deviation over $5$ different runs with different random seeds.}
    \label{fig:bayes4cl:protocl_mem_ablation}
\end{figure}

\noindent \textbf{Preventing forgetting.} We use coresets to retain the class prototypes. Coresets are randomly sampled data from previous tasks which are then stored together in a replay buffer and added to the next task training set. At the end of learning a task $\mathcal{T}_t$, we retain a subset $\mathcal{M}_t \subset \mathcal{D}_t$ and augment each new task dataset to ensure that posterior parameters $\eta_t$ and prototypes are able to retain previous task information. With no coreset the average accuracy over Split-MNIST is $33.25 \pm 0.15$,~\cref{fig:bayes4cl:protocl_mem_ablation} dramatically below $93.73 \pm 1.05$ from~\cref{tab:bayes4cl:proto_cl_res}. So the main mechanism for preventing forgetting is the replay buffer which enables the network to maintain a prototype per class, rather than the sequential Bayesian inference in the prototype embedding space similarly to functional Bayesian regularization methods~\cite{titsias2019functional}.

\noindent \textbf{Class-incremental learning.} In this CL setting we do not tell the CL agent which task it is being evaluated on with a task identifier $t$. So we cannot use the task identifier to select a specific head to use for classifying a test point, for example. Also, we require the CL agent to identify each class, $\{0, \ldots, 9\}$ for Split-MNIST and Split-CIFAR10 for example, and not just $\{0, 1\}$ as in domain-incremental learning. Class-incremental learning is more general, realistic, and harder a problem setting and thus important to focus on rather than other settings, despite domain-incremental learning also being compatible with sequential Bayesian inference as described in~\cref{eq:bayes_cl}.

\setlength{\tabcolsep}{17.5pt}
\begin{table*}[hbtp]
  \centering
  \caption{Mean accuracies across all tasks over CL vision benchmarks for \emph{class incremental learning}~\citep{van2019three}. All results are averages and standard errors over $10$ seeds. $^{*}$Uses the predictive entropy to make a decision about which head for class incremental learning. Training times have been benchmarked using an Nvidia RTX3090 GPU.}
  \label{tab:bayes4cl:proto_cl_cifar}
  \resizebox{1.0\linewidth}{!}{
  \begin{tabular}{lccc}
  \toprule
  \bfseries Method & \bfseries Training time (sec) $(\downarrow)$ & \bfseries Split CIFAR-10 (acc) $(\uparrow)$  \\
  \midrule
  FROMP~\citep{pan2020continual} & $1425 \pm 28$ & \hspace*{5pt}$48.92 \pm 10.86$  \\
  S-FSVI~\citep{Rudner2021cfsvi} & $44434 \pm 91$ & $50.85 \pm 3.87$ \\
  ProtoCL (\textbf{ours}) & $\bm{384 \pm 6}$ & \bm{$55.81 \pm 2.10$} \\
  \midrule
  & & \bfseries Split CIFAR-100 (acc) \\
  \midrule
  S-FSVI~\citep{Rudner2021cfsvi} & $37355 \pm 1135$ & $20.04 \pm 2.37$ \\
  ProtoCL (\textbf{ours}) & $\bm{1425 \pm 28}$ & $\bm{23.96 \pm 1.34}$ \\
  \bottomrule
  \end{tabular}
  }
\end{table*}

\noindent \textbf{Implementation.} For Split-MNIST and Split-FMNIST the baselines and ProtoCL all use two-layer NNs with a hidden state size of 200. For Split-CIFAR10 and Split-CIFAR100, the baselines and ProtoCL use a four-layer convolution neural network with two fully connected layers of size $512$ similarly to~\citet{pan2020continual}. For ProtoCL and all baselines that rely on replay, we fix the size of the coreset to $200$ points per task. For all ProtoCL models, we allow the prior Dirichlet parameters to be learned and set their initial value to $0.7$ found by a random search over MNIST with ProtoCL. An important hyperparameter for ProtoCL is the embedding dimension of the Gaussian prototypes for Split-MNIST and Split-FMNIST this was set to $128$ while for the larger vision datasets, this was set to $32$ found using grid-search.

\noindent \textbf{Results.} ProtoCL produces good results on CL benchmarks on par or better than S-FSVI~\citep{Rudner2021cfsvi} which is state-of-the-art among Bayesian CL methods while being a lot more efficient to train and without requiring expensive variational inference. ProtoCL can flexibly scale to larger CL vision benchmarks producing better results than S-FSVI. Code to reproduce all experiments can be found here \textcolor{purple}{\href{https://github.com/skezle/bayes_cl_posterior}{https://github.com/skezle/bayes\_cl\_posterior}}. All our experiments are in the more realistic class incremental learning setting, which is a harder setting than those reported in most CL papers, so the results in~\cref{tab:bayes4cl:proto_cl_res} are lower for certain baselines than in the respective papers. We use $200$ data points per task, see~\cref{fig:bayes4cl:protocl_mem_ablation} for a sensitivity analysis of the performance over the Split-MNIST benchmark as a function of core size for ProtoCL.

The stated aim of ProtoCL is not to provide a novel state-of-the-art method for CL, but rather to propose a simple baseline that takes an alternative route than weight-space sequential Bayesian inference. We can achieve strong results that mitigate forgetting, namely by modeling the generative CL process and using sequential Bayesian inference over a few parameters in the class prototype embedding space. We argue that modeling the generative CL process is a fruitful direction for further research rather than attempting sequential Bayesian inference over the weights of a BNN. ProtoCL scales to $10$ tasks of Split-CIFAR100 which to the best of our knowledge, is the most number of tasks and classes which has been considered by previous Bayesian continual learning methods.

\section{Discussion and Conclusions}
\label{sec:conc}
In this paper, we revisited the use of sequential Bayesian inference for CL. We can use sequential Bayes to recursively build up the multi-task posterior~Equation~\eqref{eq:bayes_cl}. Previous methods have relied on approximate inference and see little benefit over SGD. We test the hypothesis of whether this poor performance is due to the approximate inference scheme by using HMC in two simple CL problems. HMC asymptotically samples from the true posterior, and we use a density estimator over HMC samples to use as a prior for a new task within the HMC sampling process. We perform many checks for HMC convergence. This density is multi-modal and accurate with respect to the current task but is not able to improve over using an approximate posterior. This demonstrates just how challenging it is to work with BNN weight posteriors. The~source of error comes from the density estimation step. We then look at an analytical example of sequential Bayesian inference where we perform exact inference; however, due to model misspecification, we observe forgetting. The~only way to ensure a well-specified model is to assess the multi-task performance over all tasks a priori. This might not be possible in online CL settings. We then model an analytical example over Bayesian NNs and, under certain assumptions, show that if there are task data imbalances then this will cause forgetting; data imbalances are a common problem in many areas of machine learning in addition to continual learning~\cite{chrysakis2020online}. Sequential Bayesian inference is also not exempt from the effects of task data imbalances. Because~of these results, we argue against performing weight space sequential Bayesian inference and instead model the generative CL problem. We introduce a simple baseline called ProtoCL. ProtoCL does not require complex variational optimization and achieves competitive results to the state-of-the-art in the realistic setting of class incremental~learning.

This conclusion should not be a surprise since the latest Bayesian CL papers have all relied on multi-head architectures or inducing points/coresets to prevent forgetting, rather than better weight-space inference schemes. Our observations are in line with recent theory from~\citep{knoblauch2020optimal}, which states that optimal CL requires perfect memory. Although~the results were shown with deterministic NNs the same results follow for BNN with a single set of parameters. Future research efforts should focus on more functional approaches to sequential Bayesian inference, in which previous task functions are remembered~\citep{titsias2019functional, pan2020continual, Rudner2022fsvi}. This shifts the problem of remembering previous task functions to a coreset similar to sparse variational-inference Gaussian Processes~\citep{titsias2009variational, hensman2013gaussian}.

\vspace{6pt}

\authorcontributions{
S.K lead the research including conceptualization, performing the experiments and writing the paper. S.J.R helped with conceptualization. A.C helped helped with the development of the ideas and the implementation of HMC with a density estimator as a prior. T.G.J.R ran the S-FSVI baselines for the class incremental continual learning experiments. T.G.J.R, A.C and S.J.R helped to write the paper.}

\funding{S.K. acknowledges funding from the Oxford-Man Institute of Quantitative Finance. T.G.J.R. acknowledges funding from the Rhodes Trust, Qualcomm, and~the Engineering and Physical Sciences Research Council (EPSRC). This material is based upon work supported by the United States Air Force and DARPA under Contract No. FA8750-20-C-0002. Any opinions, findings and conclusions or recommendations expressed in this material are those of the author(s) and do not necessarily reflect the views of the United States Air Force and DARPA.}

\institutionalreview{Not applicable.}

\dataavailability{All data is publically available, code to reproduce all experiments can be found here \url{https://github.com/skezle/bayes_cl_posterior}.}

\acknowledgments{We would like to thank Sebastian Farquhar, Laurence Aitchison, Jeremias Knoblauch, and~Chris Holmes for discussions. We would also like to thank Philip Ball for his help with writing the~paper.}

\conflictsofinterest{The authors declare no conflict of interest. The~funders had no role in the design of the study; in the collection, analyses, or~interpretation of data; in the writing of the manuscript; or in the decision to publish the~results.}
\newpage
\abbreviations{Abbreviations}{
The following abbreviations are used in this manuscript:\\

\noindent 
\begin{tabular}{@{}ll}
CL & Continual Learning \\
NN & Neural Network \\
BNN & Bayesian Neural Network \\
HMC & Hamiltonian Monte Carlo \\
VCL & Variational Continual Learning \\
SGD & Stochastic Gradient Descent \\
SH & Single Head \\
MH & Multi-head \\
GMM & Gaussian Mixture Model \\
ProtoCL & Prototypical Bayesian Continual Learning \\
\end{tabular}
}

\appendixtitles{yes} 
\appendixstart
\appendix

\crefalias{section}{appsec}
\crefalias{subsection}{appsec}
\crefalias{subsubsection}{appsec}


%

%
%
\newpage 
\section*{\large Appendix}
\label{sec:appendix}

\section*{Table of Contents}
\vspace*{-10pt}
\startcontents[sections]
\printcontents[sections]{l}{1}{\setcounter{tocdepth}{2}} 

\section{Continual Learning Experimental Scenarios}
\label{sec:cl_scenarios}
To evaluate different continual learning methods, we need to define the commonly used continual learning scenarios that are used in the literature and throughout this paper.

Previous work has introduced models for continual learning where the NN architectures used a feature extractor shared among all continual learning tasks, but a bespoke feature to output linear layer is trained for each task and then frozen~\cite{zenke2017continual}. Alternatively, other works have sought methods that do not require this manual selection of different feature to output linear heads and proposed a single feature to output linear layer which is shared among all tasks in continual learning~\cite{farquhar2018towards}. In this section, we will categorize and systematically interpret the different continual learning scenarios, similar to previous important work~\cite{hsu2018re, van2019three}.

In terms of notation, a task $\mathcal{T}_t$ can be characterized by the conditional and marginal data distributions $p_t(y | \vx)$ and $p_t(\vx)$ and a task identifier $t$ we denote samples from the distributions $y \sim p_t(y)$ and $x \sim p_t(\vx)$.

\textbf{Task incremental learning.} This first scenario and generally the easiest scenario for continual learning. Each task is a subset of classes in the dataset where the input domains are disjoint $p_1(\vx) \neq p_2(\vx)$ while the output spaces are shared among all tasks $p_1(y) = p_2(y)$ a task identifier is also available $t$. For example, for Split CIFAR10 all tasks are a binary classification problems, and the classes are all mapped to $\{0, 1\}$ for each task. The task identifier can be used to select a different linear layer per task~\cite{Kirkpatrick, nguyen2018variational, zenke2017continual} --- described as a multi-head network. The task identifier can be used during training and for evaluation. 

\textbf{Domain incremental learning.} In this scenario the domain explicitly increases, since $p_1(\vx) \neq p_2(\vx)$ but the learner is required to retain knowledge about previous domains. The output spaces remain shared by all tasks $p_1(y) = p_2(y)$. In contrast to task incremental learning, no task identifier is available to the continual learning agent.

\begin{figure}
    \centering
    \includegraphics[width=1.00\textwidth]{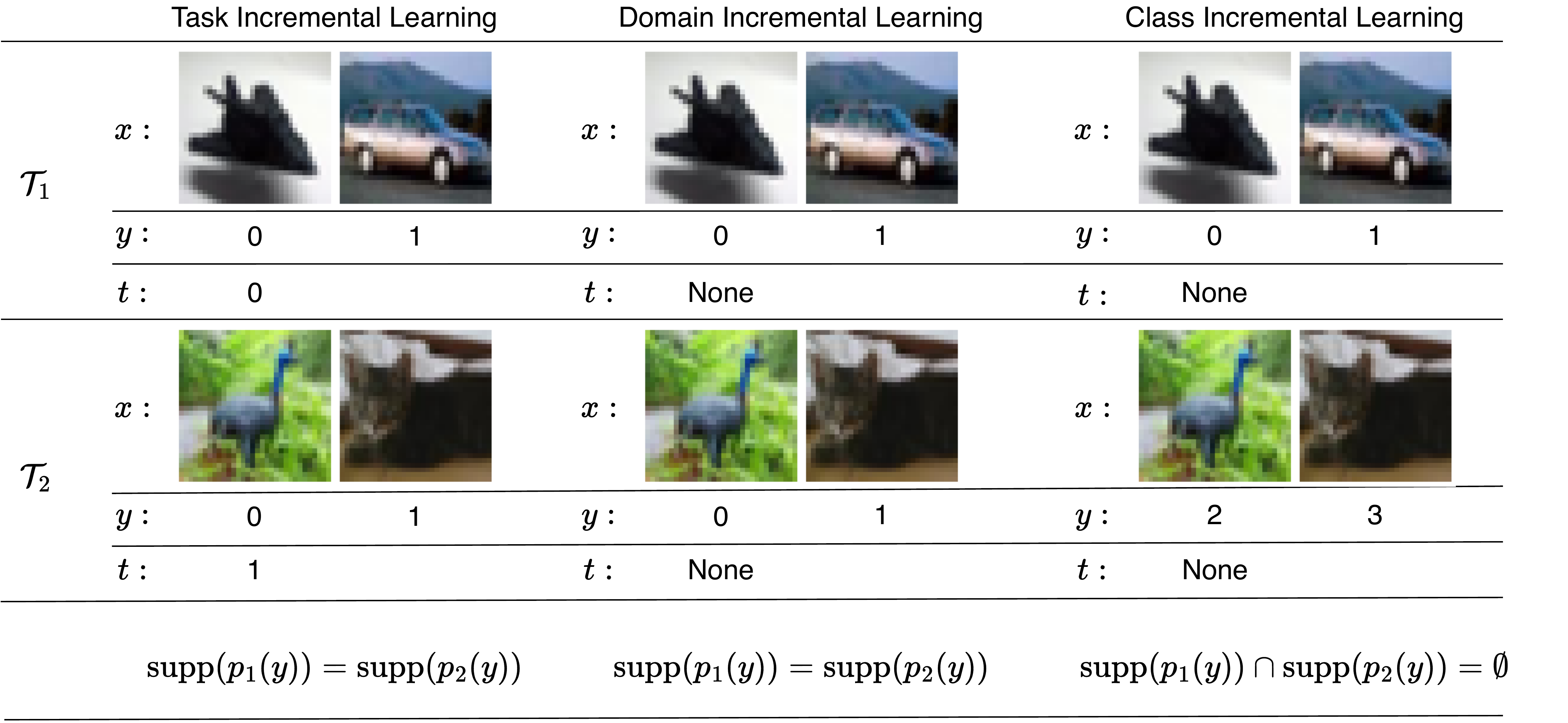}
    \caption{Three continual learning scenarios. Example datapoints from $2$ tasks from the Split CIFAR10 benchmark. The first task is binary classification of airplanes versus automobiles and the second task is binary classification of birds versus cats. In each row we have a different task, in each sub-row in each task the exact class $y$ which needs to be predicted is enumerated, and the task identifier, $t$, is shown. The support of the discrete class labels is defined as $\text{supp}(P(y)) = \{y \in \{0, \ldots, 9\}: P(y) > 0\}$.}
    \label{fig:cl_scenarios}
\end{figure}

\textbf{Class incremental learning.} In this scenario the domain increases $p_1(\vx) \neq p_2(\vx)$ as the number of tasks increases,  while the number of classes seen increases as the number of tasks increases, so $p_1(y) \neq p_2(y)$. Additionally, no task identifier is available to the agent. An example is illustrated in~\cref{fig:cl_scenarios}.

\textbf{Multi-head versus single-head networks.} A common design choice not exclusively used in continual learning is to have an output linear layer per task, or a linear head per task, $h$, map to outputs $h: \mathcal{Z} \rightarrow \mathcal{Y}$, where $y \in \mathcal{Y}$. Such that a continual learning agent uses a separate head per task $\left\{h_i\right\}_{i=1}^T$, these methods are called multi-headed while those that use one head are called single-headed. Note that the multi-head networks are only compatible with \emph{task-incremental} learning since they require knowledge of a task identifier, $t$, to select a new head during training and the head that corresponds to a specific task during evaluation. On the other hand, the single-headed network doesn't require knowledge of the task identifier during training and evaluation and so is compatible with \emph{domain-incremental} and \emph{class-incremental} learning.

\section{Continual Learning Metrics}
\label{sec:metrics}

I will define the metrics which are used to measure the performance of supervised continual learning agents and which are used in the experimental setups in this paper.

\textbf{Average Performance.} Let $p_{k, j}$ be the performance of the continual learning agent, higher performance is better, such as the accuracy on the test set $j$ after training incrementally on tasks $1, \ldots, k$ and $j \leq k$. Then the average performance at task $k$ is defined as:
\begin{align}
    \label{eq:metrics:av_performance}
    P_k = \frac{1}{k} \sum_{i=1}^{k} p_{k, i}. 
\end{align}
The higher the average performance $P_k$ over all tasks $1, \ldots, j$ the better the continual learner is at learning all tasks $\mathcal{T}_{:j}$ seen so far. For classification the performance $p_{k, j}$ is the accuracy $a_{k, j}$ and for a canonical benchmark like Split MNIST which has $5$ tasks in total then we measure the performance at the end of training on the last task i.e. $k=5$:
\begin{align}
    P_5 = \frac{1}{5} \sum_{i=1}^5 a_{5, i}
\end{align}
which averages the performance over the test sets $j = 1,\ldots, 5$. This is the main metric which is used throughout the paper as it simultaneously measures how well a continual learner is able to perform a specific task and how well it retains knowledge and prevents catastrophic forgetting.

\textbf{Average Forgetting.} This is defined as the difference between the performance after a task is trained and the current performance. This difference defines the performance gap and therefore how much the model has forgotten how to perform a task. The forgetting for the task $j$ after learning $k$ tasks $1, \ldots, k$ and $j < k$ is defined as:
\begin{align}
    \label{eq:metrics:forgetting}
    f^{k}_{j} = p_{j, j} - p_{k, j}, \quad \forall j < k.
\end{align}
The average forgetting can only be defined for the previous $k-1$ tasks as:
\begin{align}
    F_k = \frac{1}{k-1} \sum_{j=1}^{k-1} f_{j}^{k}.
\end{align}
An $F_k$ close to zero implies little forgetting. A negative forgetting implies that the performance improves throughout continual learning and the learner can transfer knowledge from future tasks when being evaluated on previous tasks. This is a very desirable and yet rare property of a continual learner. A positive $F_k$ indicates forgetting and the performance on task $j$ degrades after learning task $k$. A note of caution: we can get no forgetting while our learner has not learned anything at all: and $p_{k, j} = 0, \, \, \forall j$ implies no forgetting $F_k = 0$ which is also undesirable.

\textbf{Performance upper bounds} An upper bound to performance is how well the continual learning model can do in comparison to a multi-task model which can learn from all tasks at once. A different upper bound on performance is the single task performance; the performance of training a single different machine learning model on each individual task, this model won't benefit from transfer available to the multi-task model.

Multi-task performance as an upper bound is natural for the computer vision continual learning benchmarks since the tasks are constructed by subsetting classes. In sequential Bayesian inference, the upper bound on performance is the multi-task posterior which the sequential posterior builds up through sequential Bayesian updates~\cref{eq:bayes_cl_t2}. As we discuss in~\cref{sec:misspecification} the Bayesian multi-task performance can sub-optimal for the continual learning problem we want to solve and the continual learning solution can outperform the multi-task solution.

\section{The Toy Gaussians~Dataset}
\label{sec:toy_gaussians}
See~Figure~\ref{fig:toy_gaussians_cl2} for a visualization of the toy Gaussians dataset, which we use as a simple CL problem. This is used for evaluating our method for propagating the true posterior by using HMC for posterior inference and then using a density estimator on HMC samples as a prior for a new task. We construct $5$, $2$-way classification problems for CL. Each $2$-way task involves classifying adjacent circles and squares~Figure~\ref{fig:toy_gaussians_cl2}. With~a $2$ layer network with $10$ neurons we obtain a test accuracy of $1.0$ for the multi-task learning of all $5$ tasks together. Hence, according to~Equation~\eqref{eq:bayes_cl} a BNN with the same size should be able to learn all $5$ binary classification tasks continually by sequentially building up the~posterior.



\section{HMC Implementation~Details}
\label{sec:hmc_impl_details}
We set the prior for $\mathcal{T}_1$, to~$p_1(\rvtheta) = \mathcal{N}(0, \tau^{-1}\mathbb{I})$ with $\tau = 10$. We burn-in the HMC chain for $1000$ steps and sample for $10,000$ more steps and run $20$ different chains to obtain samples from our posterior, which we then pass to our density estimator. We use a step size of $0.001$ and trajectory length of $L=20$, see~Appendix \ref{sec:hmc_diagnostics} for further implementation details of the density estimation procedure. For~the GMM, we optimize for the number of components by using a holdout set of HMC~samples.

\section{Density Estimation~Diagnostics}
\label{sec:hmc_diagnostics}

\begin{figure}[H]
\begin{adjustwidth}{-\extralength}{0cm}\centering
  \includegraphics[width=0.95\linewidth]{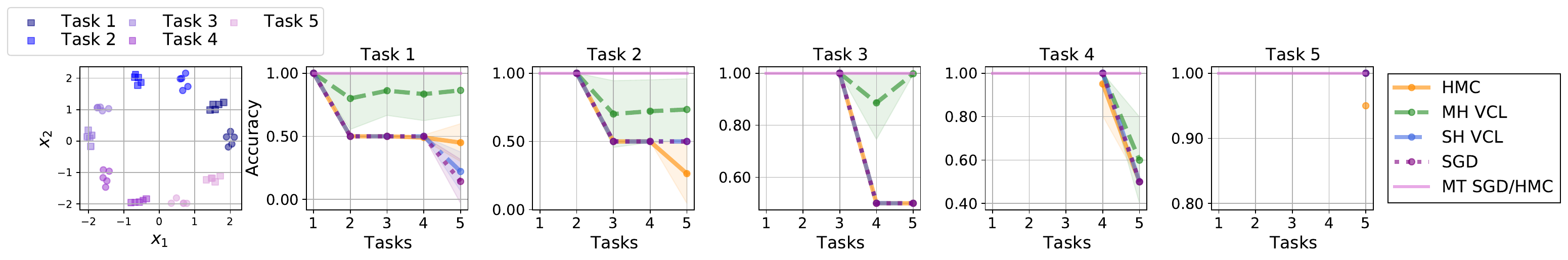}
\end{adjustwidth}
  \caption{Continual learning binary classification accuracies from the toy Gaussian dataset similar to~\citep{henning2021posterior} using $10$ random seeds. The~pink solid line is a multi-task (MT) baseline test accuracy using SGD/HMC.}
  \label{fig:toy_gaussians_cl2}
\end{figure}

\begin{figure}[H]
    \includegraphics[width=0.9\textwidth]{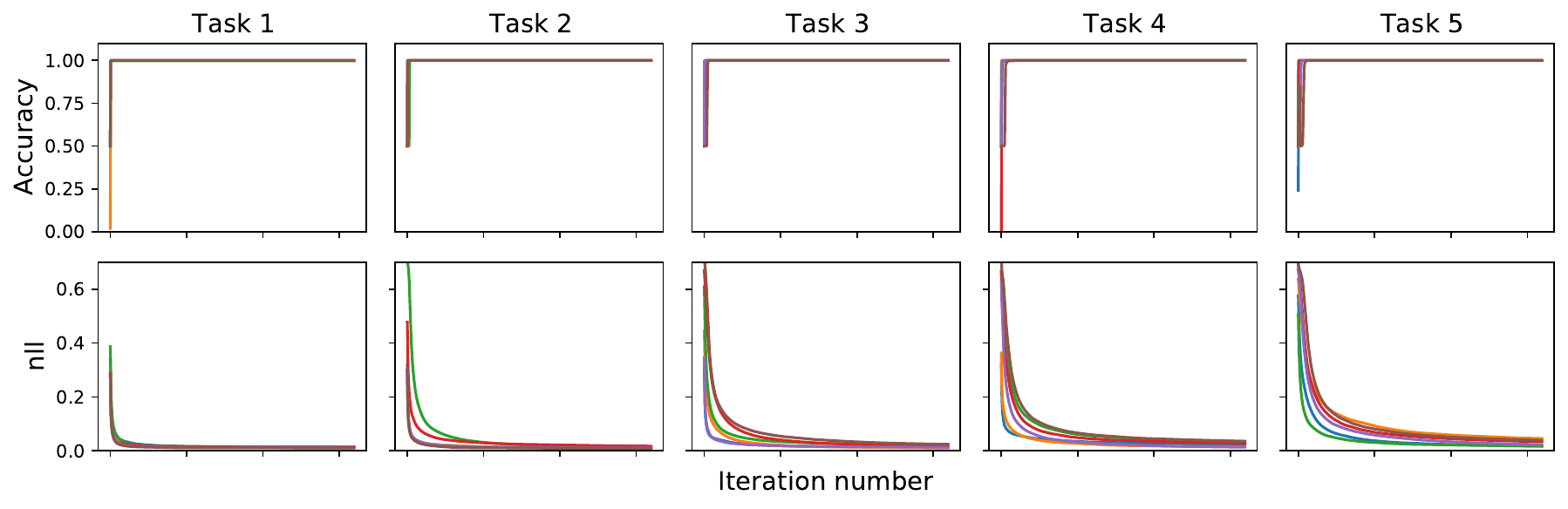}
    \caption{Convergence plots from a one randomly sampled HMC chain (of $20$) for each task over $10$ different runs (seeds) for $5$ tasks from the toy Gaussian dataset similar to~\citet{henning2021posterior} (visualized in~Figure~\ref{fig:toy_gaussians_cl2}). We use a GMM density estimator as the prior conditioned on the previous task~data.}
    \label{fig:tg_convergence_diagnostics}
\end{figure}

\begin{figure}[H]
    \includegraphics[width=0.9\textwidth]{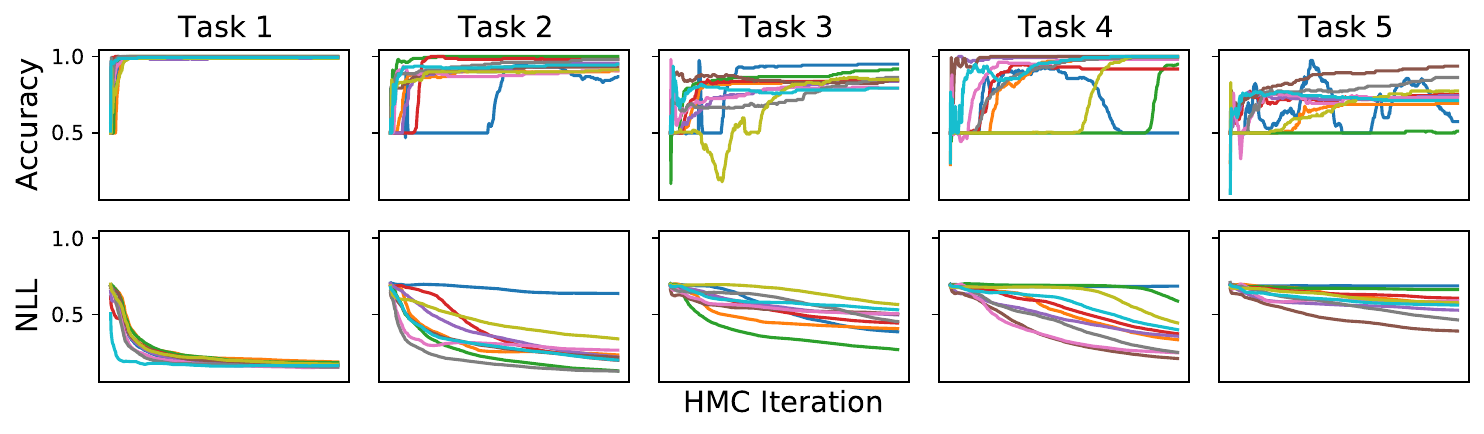}
    \caption{Convergence plots from a randomly sampled HMC chain (of $20$) for each task over $10$ different seeds for $5$ tasks from the toy dataset from~\citep{pan2020continual} (see~Figure~\ref{fig:tg_res} for a visualization of the data). We use a GMM density estimator as a~prior.}
    \label{fig:td_convergence_diagnostics}
\end{figure}

We provide plots to show that the HMC chains indeed sample from the posterior have converged in~Figures~\ref{fig:tg_convergence_diagnostics} and~\ref{fig:td_convergence_diagnostics}. We run $20$ HMC sampling chains and randomly select one chain to plot for each seed (of $10$). We run HMC over $10$ seeds and aggregate the results~Figures~\ref{fig:tg_res} and~\ref{fig:toy_gaussians_cl2}. The~posteriors $p(\theta | \mathcal{D}_1), \ldots$ are approximated with a GMM and used as a prior for the second task and so~forth.

\begin{figure}[H]
    \includegraphics[width=0.9\textwidth]{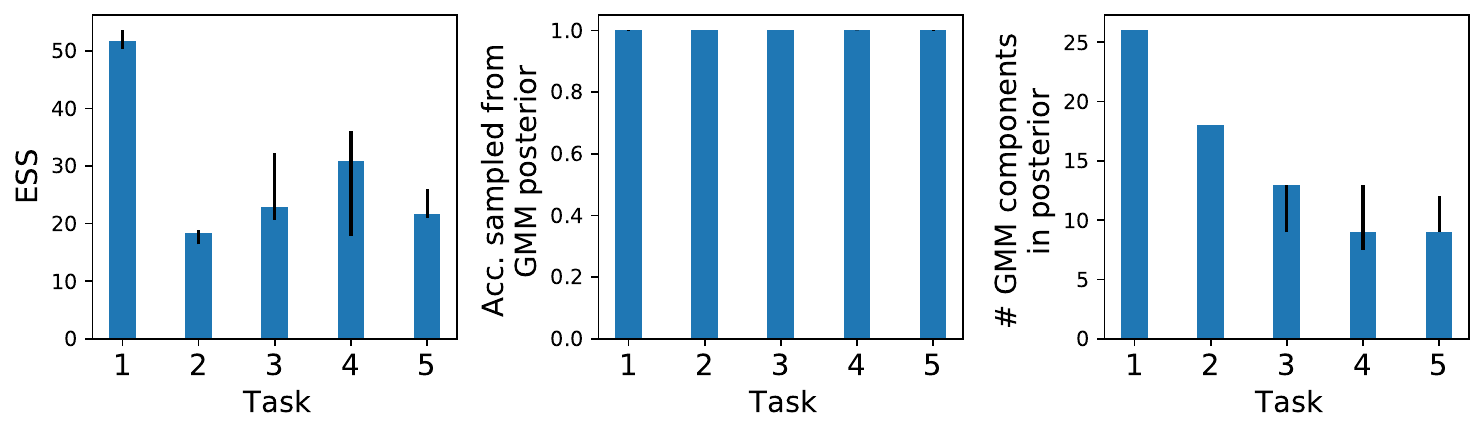}
    \caption{Diagnostics from using a GMM prior fit to samples of the posterior generated from HMC, all results are for $10$ random seeds. \textbf{Left,} effective sample sizes (ESS) of the resulting HMC chains of the posterior, all are greater than those reported in other works using HMC for BNNs~\citep{cobb2020scaling}. \textbf{Middle,} the accuracy of the BNN when using samples from the GMM density estimator instead of the samples from HMC. \textbf{Right,} The optimal number of components of each GMM posterior fitted with a holdout set of HMC samples by maximizing the~likelihood.}
    \label{fig:tg_gmm_diagnostics}
\end{figure}

\begin{figure}[H]
    \includegraphics[width=0.7\textwidth]{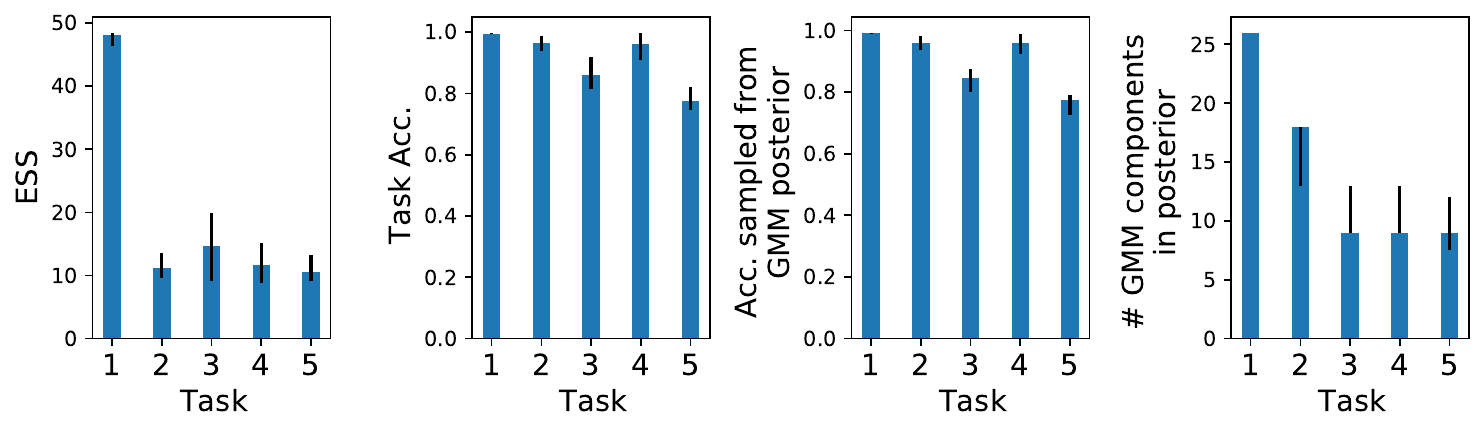}
    \caption{Diagnostics from using a GMM to fit samples of the posterior HMC samples, all results are for $10$ random seeds on the toy dataset from~\citet{pan2020continual} (and visualized in~Figure~\ref{fig:tg_res}). \textbf{Left}, effective sample sizes (ESS) of the resulting HMC chains of the posterior, all are greater than those reported in other works using HMC for BNNs~\citep{cobb2020scaling}. \textbf{Middle left}. the current task accuracy from HMC sampling. \textbf{Middle right}, the accuracy of the BNN when using samples from the GMM density estimator instead of the converged HMC samples. \textbf{Right}, The optimal number of components of each GMM posterior fitted with a holdout set of HMC samples by maximizing the~likelihood.}
    \label{fig:td_gmm_diagnostics}
\end{figure}

We provide empirical evidence to show that the density estimators have fit to HMC samples of the posterior in~Figures~\ref{fig:tg_gmm_diagnostics} and~\ref{fig:td_gmm_diagnostics}, where we show the number of components of the GMM density estimator, which we use as a prior for a new task, are all multi-modal posteriors. We show the BNN accuracy when sampling BNN weights from our GMM all recover the accuracy of the converged HMC samples. The~effective sample size (ESS) of the $20$ chains is a measure of how correlated the samples are (higher is better). The~reported ESS values for our experiments are in line with previous work which uses HMC for BNN inference~\citep{cobb2020scaling}.

\unskip
\unskip
\unskip
\unskip
\unskip

\section{Prototypical Bayesian Continual~Learning}
\label{sec:protocl_derivation}

ProtoCL models the generative process of CL where new tasks are comprised of new classes $j \in \{1, \ldots, J\}$ of a total of $J$ and can be modeled by using a categorical distribution with a Dirichlet prior:
\begin{align}
    y_{i, t} \sim \text{Cat}(p_{1:J}), \quad p_{1:J}\sim \textrm{Dir}(\alpha_t).
\end{align} 
We learn a joint embedding space for our data with a NN, $\vz = f(\vx ; \vw)$ with parameters $\vw$. The~embedding space for each class is Gaussian whose mean has a prior which is also Gaussian:
\begin{align}
    \vz_{it}|y_{it} \sim \mathcal{N}(\bar{\rvz}_{yt}, \Sigma_{\epsilon}), \quad \bar{\rvz}_{yt} \sim \mathcal{N}(\vmu_{yt}, \Lambda^{-1}_{yt}).
\end{align} 

By ensuring that we have an embedding per class and using a memory of past data, we ensure that the embedding does not drift. The~posterior parameters are $\eta_{t} = \{\alpha_{t}, \vmu_{1:J, t}, \Lambda^{-1}_{1:J, t}\}$. 

\subsection{Inference}

As the Dirichlet prior is conjugate with the categorical distribution and so is the Gaussian distribution with a Gaussian prior over the mean of the embedding, then we can calculate posteriors in closed form and update our parameters as we see new data online without using gradient-based updates. We perform gradient-based learning of the NN embedding function $f( \,\cdot \, ; \vw)$ with parameters $\vw$. We optimize the model by maximizing the log-predictive posterior of the data and use the softmax over class probabilities to perform predictions. The~posterior over class probabilities $\{p_{j}\}_{j=1}^J$ and class embeddings $ \{\bar{\vz}_{y_{j}}\}_{j=1}^J $ is denoted as $p(\theta)$ for short hand and has parameters are $\eta_{t} = \{\alpha_{t}, \vmu_{1:J, t}, \Lambda^{-1}_{1:J, t}\}$ are updated in closed form at each iteration of gradient~descent.

\subsection{Sequential~Updates} 
\label{sec:protocl_seq_updates}
We can obtain our posterior: 
\begin{align}
    p(\vtheta_t|\mathcal{D}_t) &\propto p(\mathcal{D}_t | \vtheta_t)p(\vtheta_t) \\
    &= \prod^{N_t}_{i=1}p(\vz^i_t|y^i_t; \bar{\vz}_{y_t}, \Sigma_{\epsilon, y_t}) p(y^i_t|p_{1:J})p(p_{i:J};\alpha_t)p(\bar{\vz}_{y_t}; \vmu_{y_t, t}, \Lambda^{-1}_{y_t, t}) \\
    &= \mathcal{N}(\mu_{t+1}, \Sigma_{t+1}) \textrm{Dir}(\alpha_{t+1}),
\end{align}
where $N_t$ is the number of data points seen during update $t$. Concentrating on the Categorical-Dirichlet conjugacy:
\begin{align}
    \textrm{Dir}(\alpha_{t+1}) &\propto p(p_{1:J};\alpha_t)\prod^{N_t}_{i=1}p(y^{i}_{t};p_{i:J}) \\
    &\propto \prod^{J}_{j=1}p_{j}^{\alpha_j - 1}\prod^{N_t}_{i=1}\prod^{J}_{j=1}p_j^{\mathbb{I}(y_{t}^i = j)} \\
    &= \prod^{J}_{j=1}p_j^{\alpha_j - 1 + \sum_{i=1}^{N_t}\mathbb{I}(y_t^i = j)}.
\end{align}
Thus:
\begin{align}
    \alpha_{t+1, j} = \alpha_{t, j} + \sum^{N_t}_{i=1}\mathbb{I}(y^i_{t} = j).
\end{align}
Moreover, due to Gaussian-Gaussian conjugacy, then the posterior for the Gaussian prototype of the embedding for each class is:
\begin{align}
    \mathcal{N}(\vmu_{t+1}, \Lambda_{t+1}) &\propto \prod^{N_t}_{i=1}\mathcal{N}(\vz_t^i | y^i_t; \bar{\vz}_{y_t}, \Sigma_{\epsilon}) \mathcal{N}(\bar{\vz}_{y_t}; \vmu_{y_t, t}, \Lambda^{-1}_{y_t}) \\
    &= \prod_{y_t \in \{1, \ldots, J\}} \mathcal{N}(\vz_{y_t} | y_t; \bar{\vz}_{y_t}, \frac{1}{N_{y_t}}\Sigma_{\epsilon}) \mathcal{N}(\bar{\vz}_{y_t}; \vmu_{y_{t+1}}, \Lambda^{-1}_{y_t}) \\
    &= \prod_{y_{t} \in \{1, \ldots, J\}} \mathcal{N}(\bar{\vz}_{y_t};\vmu_{t+1}, \Lambda^{-1}_{y_{t+1}}),
\end{align}
where $N_{y_t}$ is the number of points of class $y_t$ from the set of all classes $C = \{1, \ldots, J\}$. The~update equations for the mean and variance of the posterior are:
\begin{align}
    \Lambda_{y_{t+1}} &= \Lambda_{y_t} + N_{y_t} \Sigma^{-1}_{\epsilon}, \quad \forall y_t \in C_t\\
    \Lambda_{y_{t+1}}\vmu_{y_{t+1}} &= N_{y_{t}} \Sigma^{-1}_{\epsilon} \bar{\vz}_{y_t} + \Lambda_{y_t} \vmu_{y_t}, \quad \forall y_t \in C_t.
\end{align}

\subsection{ProtoCL~Objective} 
\label{sec:protocl_obj}
The posterior predictive distribution we want to optimize is:
\begin{equation}
    p(\vz, y) = \int p(\vz, y | \rvtheta ; \eta)p(\rvtheta ; \eta)d\rvtheta,
\end{equation}
where $p(\rvtheta)$ denotes the distributions over class probabilities $\{p_{j} \}_{j=1}^J$ and mean embeddings $\{\bar{\vz}_{y_j}\}_{j=1}^J$,
\begingroup
\makeatletter\def\f@size{9}\check@mathfonts
\def\maketag@@@#1{\hbox{\m@th\normalsize\normalfont#1}}
\begin{align}
    p(\vz, y) &= \int \prod^{N_t}_{i=1}p(\vz_{it}| y_{it}; \bar{\vz}_{y_t}, \Sigma_{\epsilon}) p(y_{it}|p_{1:J})p(p_{1:J} ; \alpha_t)p(\bar{\vz}_{y_t}; \vmu_{y_{t}, t}, \Lambda^{-1}_{y_t, t})d p_{1:J} d\bar{\vz}_{y_t} \\
    &= \int \prod^{N_t}_{i=1} p(\vz_{it} | y_{it}; \vz_{y_t}, \Sigma_{\epsilon}) p(\bar{\vz}_{y_t}; \vmu_{y_{t}, t}, \Lambda^{-1}_{y_t, t}) d\bar{\vz}_{y_t}\underbrace{\int \prod^{N_t}_{i=1} p(y_{it} | p_{1:J}) p(p_{1:J} ; \alpha_t) d p_{1:J}}_{\prod_i p(y_i) = p(y)} \\
    &= p(y) \prod^{N_t}_{i=1} Z_{i}^{-1}  \int \mathcal{N}(\bar{\vz}_{y_{it}}; \vc, C)d\bar{\vz}_{y_t} \label{eq:matrix_cook_book}\\
    &= p(y) \prod^{N_t}_{i=1} \mathcal{N}(\vz_{it} | y_{it} ; \vmu_{y_t, t}, \Sigma_{\epsilon} + \Lambda^{-1}_{y_t, t}).
\end{align}
\endgroup
where in Equation~(\ref{eq:matrix_cook_book}) we use §8.1.8 in~\citep{petersen2008matrix}. The~term $p(y)$ is:
\begin{align}
    p(y) &= \int p(y|p_{1:J})p(p_{1:J};\alpha_t) dp_{1:J} \\
    &=\int p_y \frac{\Gamma(\sum^{J}_{j=1} \alpha_j)}{\prod^{J}_{j=1}\Gamma(\alpha_j)} \prod^J_{j=1} p_j^{\alpha_j-1} dp_{1:J} \\
    &= \frac{\Gamma(\sum^{J}_{j=1} \alpha_j)}{\prod^{J}_{j=1}\Gamma(\alpha_j)} \int \prod^J_{j=1} p_j^{\mathbb{I}(y=j) + \alpha_j - 1} dp_{1:J} \\
    &= \frac{\Gamma(\sum^{J}_{j=1} \alpha_j)}{\prod^{J}_{j=1}\Gamma(\alpha_j)} \frac{\prod^{J}_{j=1} \Gamma(\mathbb{I}(y=j) + \alpha_j)}{\Gamma(1 + \sum^{J}_{j=1} \alpha_j )} \\
    &= \frac{\cancel{\Gamma(\sum^{J}_{j=1} \alpha_j)}}{\prod^{J}_{j=1}\Gamma(\alpha_j)} \frac{\prod^{J}_{j=1} \Gamma(\mathbb{I}(y=j) + \alpha_j)}{\sum^{J}_{j=1} \alpha_j \cancel{\Gamma(\sum^{J}_{j=1} \alpha_j )}} \\
    &= \frac{\prod^{J}_{j=1, j\neq y}\Gamma(\alpha_j)}{\prod^{J}_{j=1}\Gamma(\alpha_j)} \frac{\Gamma(1+\alpha_y)}{\sum^{J}_{j=1} \alpha_j } \\
    &= \frac{\prod^{J}_{j=1, j\neq y}\Gamma(\alpha_j)}{\prod^{J}_{j=1}\Gamma(\alpha_j)} \frac{\alpha_y\Gamma(\alpha_y)}{\sum^{J}_{j=1} \alpha_j } \\
    &= \frac{\alpha_y}{\sum^{J}_{j=1} \alpha_j},
\end{align}
where we use the identity $\Gamma(n+1) = n\Gamma(n)$.

\begin{figure}[H]
    \includegraphics[width=0.6\textwidth]{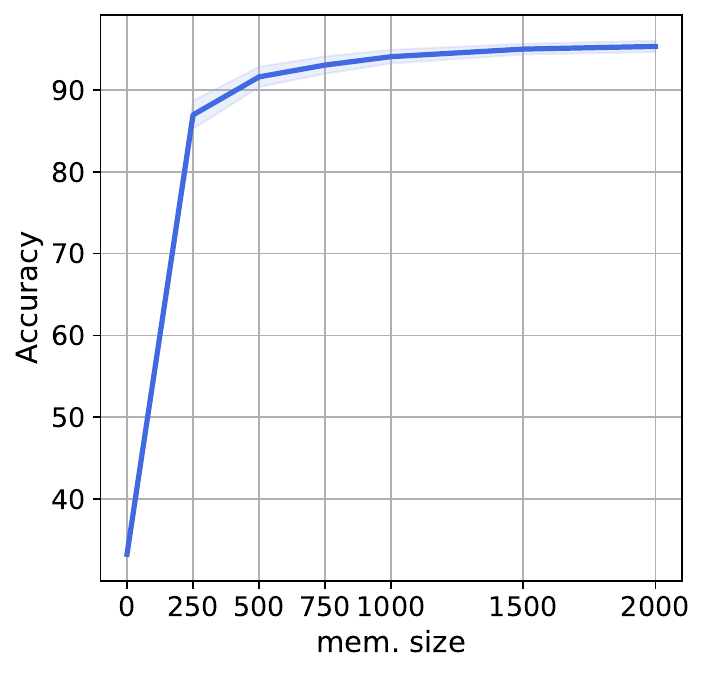}
\caption{Split-MNIST average test accuracy over five tasks for different memory sizes. On~the x-axis, we show the size of the entire memory buffer shared by all five tasks. Accuracies are over a mean and standard deviation over five different runs with different random~seeds.}
    \label{fig:protocl_mem_ablation}
\end{figure}
\unskip

\begin{figure}[H]
    \includegraphics[width=0.9\textwidth]{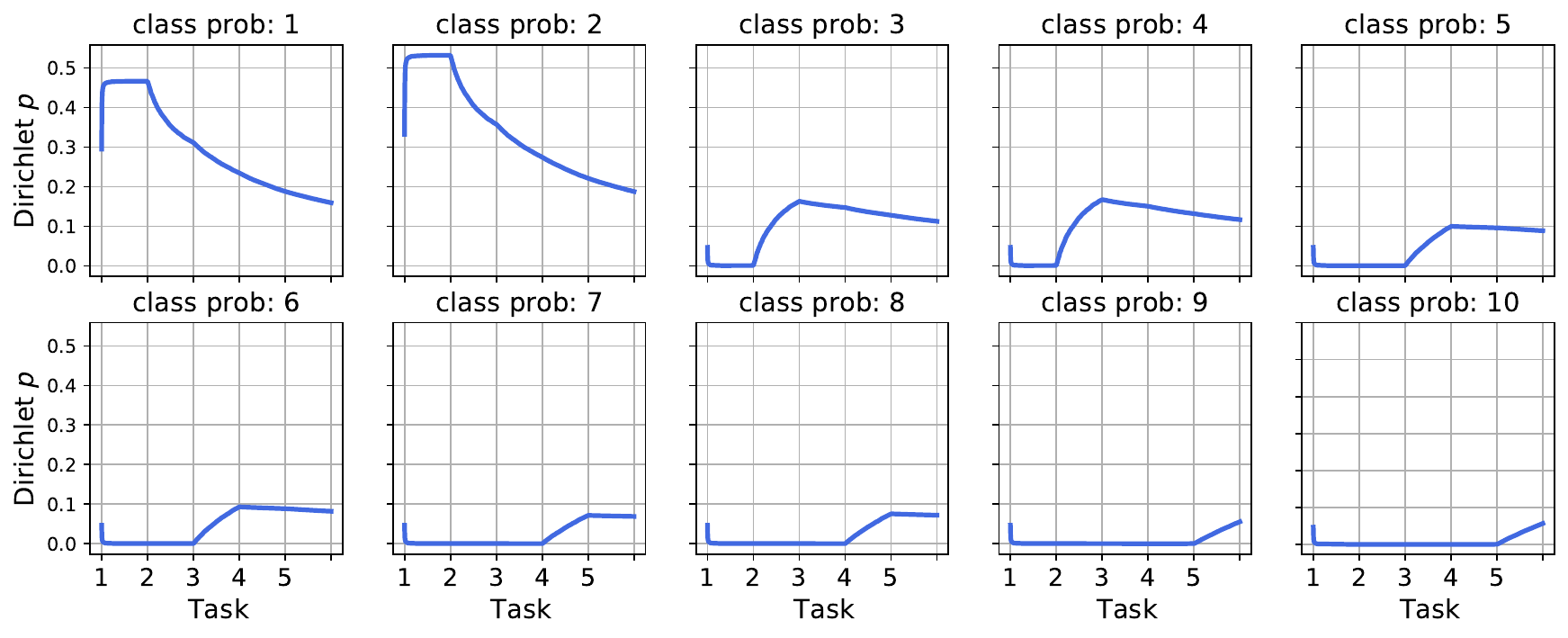}
    \caption{The evolution of the Dirichlet parameters $\alpha_t$ for each class in Split-MNIST tasks for ProtoCL. All $\alpha_{j}$ are shown over $10$ seeds with $\pm 1$ standard error. By~the end of training, all classes are roughly equally likely, as~we have trained on equal amounts of all~classes.}
    \label{fig:protocl_dir_ablation}
\end{figure}
\unskip

\subsection{Predictions} 
\label{sec:proto_cl_predictions}
To make a prediction for a test point $\vx^*$:
\begin{align}
    p(y^*=j | \vx^*, \vx_{1:t}, y_{1:t}) &= p(y^*=j | \vz^*, \rvtheta_t) \\
    &= \frac{p(\vz^*|y^*=j, \rvtheta_t)p(y^*=j|\rvtheta_t)}{\sum_i p(\vz^*|y^*=i, \eta_t)p(y^*=i|\rvtheta_t)} \\
    &= \frac{p(y^* =j , \vz^* | \rvtheta_t)}{\sum_{i}p(y=i, \vz^* | \rvtheta_t)},
\end{align}
where $\rvtheta_t$ are sufficient statistics for $(\vx_{1:t}, y_{1:t})$.

\noindent \textbf{Preventing \ forgetting.} As we wish to retain the task-specific prototypes, at~the end of learning a task $\mathcal{T}_t$ we take a small subset of the data as a memory to ensure that posterior parameters and prototypes do not drift, see~Algorithm~\ref{alg:protocl}.

\subsection{Experimental~Setup}
\label{sec:cl3}
The prototype variance, $\Sigma_{\epsilon}$ is set to a diagonal matrix with the variances of each prototype set to $0.05$. The~prototype prior precisions, $\Lambda_{yt}$, are also diagonals and initialized randomly and exponentiated to ensure a positive semi-definite covariance for the sequential updates. The~parameters $\alpha_j \, \forall j$ are set to $0.78$, which was found by random search over the validation set on MNIST. We also allow $\alpha_j$ to be learned in the gradient update step in addition to the sequential update step (lines 4 and 5~Algorithm~\ref{alg:protocl}), see~Figure~\ref{fig:protocl_dir_ablation} to see the evolution of the $\alpha_j$ or all classes $j$ over the course of learning~Split-MNIST.

For the Split-MNIST and Split-FMNIST benchmarks, we use an NN with two layers of size $200$ and trained for $50$ epochs with an Adam optimizer. We perform  a grid-search over learning rates, dropout rates, and weight decay coefficients. The~embedding dimension is set to $128$. For~the Split-CIFAR10 and Split-CIFAR100 benchmarks, we use the same network as~\citet{pan2020continual}, which consists of four convolution layers and two linear layers. We train the networks for $80$ epochs for each task with the Adam optimizer with a learning rate of $1 \times 10^{-3}$. 
 The~embedding dimension is set to $32$. All experiments are run on a single GPU NVIDIA RTX~3090.

\section{Sequential Bayesian Estimation as Bayesian Neural Network~Optimization}
\label{sec:bayes_as_kf_appendix}

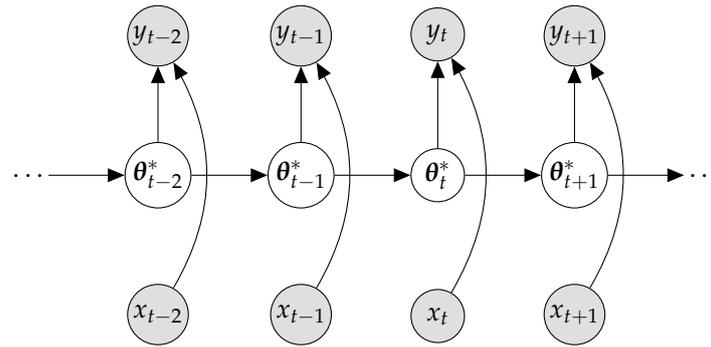
\begin{figure}[H]
     \begin{tikzpicture}
      \node[latent]                            (wtm1) {$\rvtheta^*_{t-1}$};
      \node[latent, left=of wtm1]              (wtm2) {$\rvtheta^*_{t-2}$};
      \node[latent, right=of wtm1]              (wt) {$\rvtheta^*_{t}$};
      \node[latent, right=of wt]                (wtp1) {$\rvtheta^*_{t+1}$};
      \node[obs, above=of wtm1]                 (ytm1) {$y_{t-1}$};
      \node[obs, above=of wtm2]                 (ytm2) {$y_{t-2}$};
      \node[obs, above=of wt, yshift=0.15cm]                 (yt) {$y_{t}$};
      \node[obs, above=of wtp1]               (ytp1) {$y_{t+1}$};
      \node[obs, below=of wtm1]  (xtm1) {$x_{t-1}$};
      \node[obs, below=of wtm2]  (xtm2) {$x_{t-2}$};
      \node[obs, below=of wt, yshift=-0.15cm]    (xt) {$x_{t}$};
      \node[obs, below=of wtp1]  (xtp1) {$x_{t+1}$};
      
      \node[const, left=of wtm2]  (wtm3)  {$\, \ldots \,$}; %
      \node[const, right=of wtp1]  (wtp2)  {$\, \ldots \,$}; %
      
      \edge {wtm3} {wtm2} ; %
      \edge {wtm2} {wtm1} ; %
      \edge {wtm1} {wt} ; %
      \edge {wt} {wtp1} ; %
      \edge {wtp1} {wtp2} ; %
      
      \edge {wtm2} {ytm2} ; %
      \edge {wtm1} {ytm1} ; %
      \edge {wt} {yt} ; %
      \edge {wtp1} {ytp1} ; %
      
      \path (xtm2) edge [bend right, ->]  (ytm2) ;
      \path (xtm1) edge [bend right, ->]  (ytm1) ;
      \path (xt) edge [bend right, ->]  (yt) ;
      \path (xtp1) edge [bend right, ->]  (ytp1) ;

      
    \end{tikzpicture}
     \caption{Graphical model of under which we perform inference in~\Cref{sec:data_imbalance}. Grey nodes are observed and white are latent~variables.} 
\label{fig:bnn_kf_gm}
\end{figure}

We shall consider inference in the graphical model depicted in~Figure~\ref{fig:bnn_kf_gm}. The~aim is to infer the optimal BNN weights, $\rvtheta^*_t$ at time $t$ given observations and the previous BNN weights. We assume a Gaussian posterior over weights with full covariance; hence, we model interactions between all weights. We shall consider the online setting where we see one data point $(\vx_t, y_t)$ at a time and we will make no assumption as to whether the data comes from the same task or different tasks over the course of learning. 

We set up the problem of sequential Bayesian inference as a filtering problem and we leverage the work of~\citet{aitchison2020bayesian}, which casts NN optimization as Bayesian sequential inference. We make the reasonable assumption that the distribution over weights is a Gaussian with full covariance. Since reasoning about the full covariance matrix of a BNN is intractable, we instead consider the $i$-th parameter and reason about the dynamics of the optimal estimates $\theta^*_{it}$ as a function of all the other parameters $\rvtheta_{-it}$. Each weight is functionally dependent on all others. If~we had access to the full covariance of the parameters, then we could reason about the unknown optimal weights $\theta^*_{it}$ given the values of all the other weights $\rvtheta_{-it}$. However, since we do not have access to the full covariance, another approach is to reason about the dynamics of $\theta^*_{it}$ given the dynamics of $\rvtheta_{-it}$ and assume that the values of the weights are close to those of the previous timestep~\citep{jacot2018neural} and so we cast the problem as a dynamical~system.

Consider a quadratic loss of the form:
\begin{align}
    \mathcal{L}(\vx_t, y_t; \rvtheta) = \mathcal{L}_t(\rvtheta) = -\frac{1}{2}\rvtheta^{\top} \mH \rvtheta + \rvz^{\top}_{t} \rvtheta,
\end{align} which we can arrive at by simple Taylor expansion, where $\mH$ is the Hessian which is assumed to be constant across data points but not across the parameters $\rvtheta$. If~the BNN output takes the form $f(\vx_t; \rvtheta)$, then the derivative evaluated at $\rvtheta_t$ is $\vz_t = \frac{\partial \mathcal{L}(\vx_t, y_t;\rvtheta)}{\partial \rvtheta} |_{\rvtheta = \rvtheta_t}$. 
To construct the dynamical equations for our weights, consider the gradient for a single datapoint:
\begin{align}
    \frac{\partial\mathcal{L}_t(\rvtheta)}{\partial \rvtheta} = -\mH \rvtheta + \rvz_t.
\end{align}
If we consider the gradient for the $i$-th weight while all other parameters are set to their current estimate:
\begin{align}
    \frac{\partial \mathcal{L}(\theta_i, \rvtheta_{-i})}{\partial \theta_i} = - H_{ii}\theta_{it} - \mH_{-ii}^{\top}\rvtheta_{-it} + z_{ti}.
\end{align}
When the gradient is set to zero we recover the optimal value for $\theta_{it}$, denoted as $\theta_{it}^*$:
\begin{align}
    \label{eq:bnn_kf_dynamics_app}
    \theta_{it}^{*} = -\frac{1}{H_{ii}} \mH_{-ii}^{\top} \rvtheta_{-it}.
\end{align}
since $z_{ti} = 0$ at the modes. The~equation above shows us that the dynamics of the optimal weight $\theta^{*}_{it}$ is dependent on all the other current values of the parameters $\rvtheta_{-it}$. That is, the~dynamics of $\theta^*_{it}$ are governed by the dynamics of the weights $\rvtheta_{-it}$. The~dynamics of $\rvtheta_{-it}$ are a complex stochastic process dependent on many different variables. Since reasoning about the dynamics is intractable, we instead assume a discretized Ornstein--Uhlenbeck process for the weights $\rvtheta_{-it}$ with reversion speed $\vartheta \in \R_{+}$ and noise variance $\eta^2_{-i}$:
\begin{align}
    p(\rvtheta_{-i,t+1} | \rvtheta_{-i ,t}) = \mathcal{N}((1-\vartheta)\rvtheta_{-it}, \eta_{-i}^2),
\end{align}
this implies that the dynamics for the optimal weight are defined by
\begin{align}
\label{eq:transition_dynamics_app}
    p(\theta^*_{i,t+1} | \theta^*_{i,t}) = \mathcal{N}((1-\vartheta) \theta^*_{it}, \eta^2),
\end{align}
where $\eta^2 = \eta^2_{-i} \mH^{\top}_{-ii} \mH_{-ii}$. This same assumption is made in~\citet{aitchison2020bayesian}. This assumes a parsimonious model of the dynamics. Together with our likelihood:
\begin{align}
    p(y_t | \vx_t; \rvtheta^{*}_{t}) = \mathcal{N}(y_t; f(\vx_t; \rvtheta^{*}_t), \sigma^2)
\end{align}
where $f(\,\cdot\,, \rvtheta)$ is a neural network prediction with weights $\rvtheta$, we can now define a linear dynamical system for the optimal weight $\theta^*_i$ by linearizing the Bayesian NN~\citep{jacot2018neural} and by using the transition dynamics in~Equation~\eqref{eq:transition_dynamics_app}. Thus, we are able to infer the posterior distribution over the optimal weights using Kalman filter-like updates~\citep{kalman1960new}. As~the dynamics and likelihood are Gaussian, then the prior and posterior are also Gaussian, for~ease of notation we drop the index $i$ such that $\theta^*_{it} = \theta^*_t$:
\begin{align}
    p(\theta^*_t | (\vx, y)_{t-1}, \ldots, (\vx, y)_1) &= \mathcal{N}(\mu_{t, \text{prior}}, \sigma^2_{t, \text{prior}}) \\
    p(\theta^*_t | (\vx, y)_t, \ldots, (\vx, y)_1) &= \mathcal{N}(\mu_{t, \text{post}}, \sigma^2_{t, \text{post}}) 
\end{align}

By using the transition dynamics and the prior we can obtain closed-form updates:
\begingroup
\makeatletter\def\f@size{8}\check@mathfonts
\def\maketag@@@#1{\hbox{\m@th\normalsize\normalfont#1}}
\begin{align}
    p(\theta^*_t| (\vx, y)_{t-1}, \ldots, (\vx, y)_1) &= \int p(\theta^*_t | \theta^*_{t-1}) p(\theta^*_{t-1} | (\vx, y)_{t-1}, \ldots, (\vx, y)_1) d\theta^*_{t-1} \\
    \mathcal{N}(\theta^*_t ; \mu_{t, \text{prior}}, \sigma^2_{t, \text{prior}}) &= \int \mathcal{N}(\theta^*_t ; (1-\vartheta)\theta^*_{t-1}, \eta^2) \mathcal{N}(\theta^*_{t-1} ;\mu_{t-1, \text{post}}, \sigma^2_{t-1, \text{post}})d\theta^*_{t-1}.
\end{align}
\endgroup
Integrating out $\theta^*_{t-1}$ we can obtain updates for the prior for the next timestep as follows:
\begin{align}
    \mu_{t, \text{prior}} &= (1-\vartheta)\mu_{t-1, post} \\
    \sigma^2_{t, \text{prior}} &= \eta^2 + (1 - \vartheta)^{-2}\sigma^{2}_{t-1, \text{post}}.
\end{align}
The updates for obtaining our posterior parameters: $\mu_{t, \text{post}}$ and $\sigma^2_{t, \text{post}}$, comes from applying Bayes' theorem:
\begin{align}
    \label{eq:bayes_kf_app}
    \log \mathcal{N}(\theta^*_t; \mu_{t, \text{post}}, \sigma^2_{t, \text{post}}) &\propto \log \mathcal{N}(y_t; f(\vx_t; \theta^{*}_t), \sigma^2) + \log \mathcal{N}(\theta^*_t; \mu_{t, \text{prior}}, \sigma^2_{t, \text{prior}}),
\end{align}
by linearizing our Bayesian NN such that $f(\vx_t, \theta_0) \approx f(\vx_t, \theta_0) + \frac{\partial f(\vx_t; \theta^*_t)}{\partial \theta^*_t}(\theta^*_t - \theta_0)$ and by substituting into~Equation~\eqref{eq:bayes_kf_app} we obtain our update equation for the posterior of the mean of our BNN parameters:
\begin{align}
    -\frac{1}{2\sigma^2_{t, \text{post}}} (\theta^*_{t} - \mu_{t, \text{post}})^2 &= -\frac{1}{2\sigma^2} (y - g(\vx_t)\theta^*_{t})^2 - \frac{1}{2\sigma^2_{t, \text{prior}}} (\theta^*_{t} - \mu_{t, \text{prior}})^2
    \\
    \mu_{t, \text{post}} &= \sigma^2_{t, \text{post}} \left(\frac{\mu_{t, \text{prior}}}{\sigma^2_{t, \text{prior}}} + \frac{y}{\sigma^2}g(\vx_t)\right) \label{eq:bnn_kf_mean_app},
\end{align}
where $g(\vx_t) = \frac{\partial f(\vx_t; \theta^{*}_{t})}{\partial \theta^*_{t}}$, and~the update equation for the variance of the Gaussian posterior is:
\begin{align}
    \label{eq:bnn_kf_variance_app}
    \frac{1}{\sigma^2_{t, \text{post}}} &= \frac{g(\vx_t)^2}{\sigma^2} + \frac{1}{\sigma^2_{t, \text{prior}}}.
\end{align}
From our updated equations,~Equation~\eqref{eq:bnn_kf_mean_app} and ~Equation~\eqref{eq:bnn_kf_variance_app}, we can notice that the posterior mean depends linearly on the prior and an additional data dependent term. These equations are similar to the filtering example in~\Cref{sec:misspecification}. Therefore, under certain assumptions above, a~BNN should behave similarly. If~there exists a task data imbalance, then the data term will dominate the prior term in~Equation~\eqref{eq:bnn_kf_mean_app} and could lead to forgetting of previous~tasks.

\bibliography{main}

\end{document}